\documentclass[10pt,twocolumn]{article}

\usepackage{cvpr}
\usepackage{times}
\usepackage{epsfig}
\usepackage{graphicx}
\usepackage{amsmath}
\usepackage{amssymb}
\usepackage{multirow}
\usepackage{textcomp}

\newcommand{\figname}{Figure}
\newcommand{\tabname}{Table}
\newcommand{\secname}{Section}
\newcommand{\eqname}{Equation}

\usepackage[pagebackref=true,breaklinks=true,letterpaper=true,colorlinks,bookmarks=false]{hyperref}

\cvprfinalcopy 

\def\cvprPaperID{0350} 

\ifcvprfinal\fi
\begin{document}

\title{Learning to Compose with Professional Photographs on the Web}

\author{Yi-Ling Chen$^{1}$~~~~~~Jan Klopp$^2$~~~~~~Min Sun$^3$~~~~~~Shao-Yi Chien$^2$~~~~~~Kwan-Liu Ma$^1$\\
$^1$University of California, Davis~~~~$^2$National Taiwan University~~~~$^3$National Tsing Hua University
\\\url{https://github.com/yiling-chen/view-finding-network}
}

\maketitle

\begin{abstract}
Photo composition is an important factor affecting the aesthetics in photography.
However, it is a highly challenging task to model the aesthetic properties of good compositions due to the lack of globally applicable rules to the wide variety of photographic styles.
Inspired by the thinking process of photo taking, we formulate the photo composition problem as a view finding process which successively examines pairs of views and determines their aesthetic preferences.
We further exploit the rich professional photographs on the web to mine unlimited high-quality ranking samples and demonstrate that an aesthetics-aware deep ranking network can be trained without explicitly modeling any photographic rules.
The resulting model is simple and effective in terms of its architectural design and data sampling method.
It is also generic since it naturally learns any photographic rules implicitly encoded in professional photographs.
The experiments show that the proposed view finding network achieves state-of-the-art performance with sliding window search strategy on two image cropping datasets.
\end{abstract}

\section{Introduction}
\footnote{\textcopyright~2017. This is the authors' version of this work. It is posted here for your personal use. Not for redistribution. The definitive version was published in \emph{ACM MM '17}, \url{https://doi.org/10.1145/3123266.3123274}}\emph{``Aesthetics is a beauty that is found by a relationship between things, people and environment.''}
\begin{flushright}
Naoto Fukasawa
\end{flushright}
In the past decade, a considerable amount of research efforts have been devoted to computationally model aesthetics in photography.
Most of these methods aim to either \emph{assess} photo quality by resorting to well-established photographic rules \cite{Datta:ECCV:2006,Ke:CVPR:2006,Dhar:CVPR:2011,Luo:ICCV:2011} or even to \emph{manipulate} the image content to improve visual quality \cite{Bhattacharya:MM:2010,Liu:EG:2010,Yao:2012:IJCV,Guo:PG:2012}.
However, to model photographic aesthetics remains a very challenging task due to the lack of a complete set of programmable rules to assess photo quality.
In recent years, large-scale datasets with peer-rated aesthetic scores \cite{Murray:CVPR:2012,Kong:ECCV:2016} enable aesthetics modeling with learning based approaches \cite{Murray:CVPR:2012,Lu:MM:2014,Lu:ICCV:2015,Mai:CVPR:2016,Kong:ECCV:2016}.
However, the peer-rated aesthetic scores are subject to the bias between subjects, since comparing the aesthetic of arbitrary pairs of image is inevitably ambiguous sometimes.
To mitigate the bias, one way is to get more signal than noise by enlarging the dataset.
However, it is a daunting task to collect significantly more images with peer-rated aesthetic scores.

\begin{figure}[t]
\begin{center}
\begin{tabular}{c@{\hspace{0.1cm}}c@{\hspace{0.1cm}}c@{\hspace{0.1cm}}}
\includegraphics[height=1.7cm]{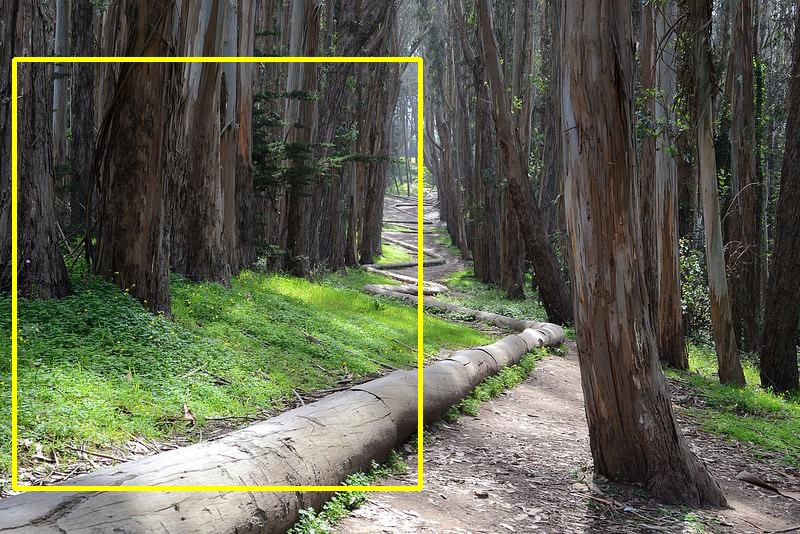}&
\includegraphics[height=1.7cm]{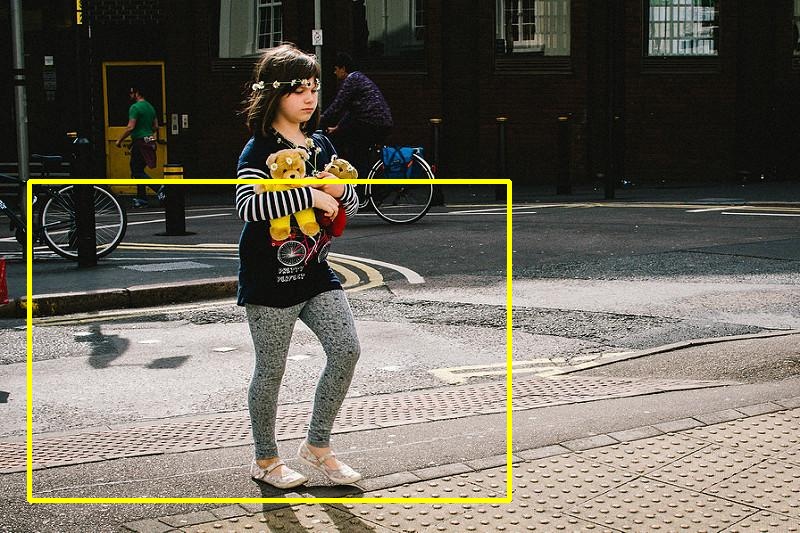}&
\includegraphics[height=1.7cm]{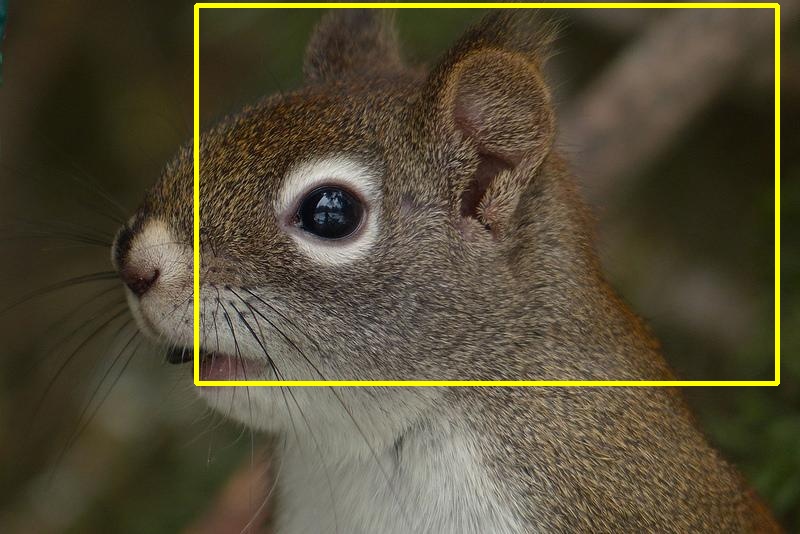}\\
\includegraphics[height=1.7cm]{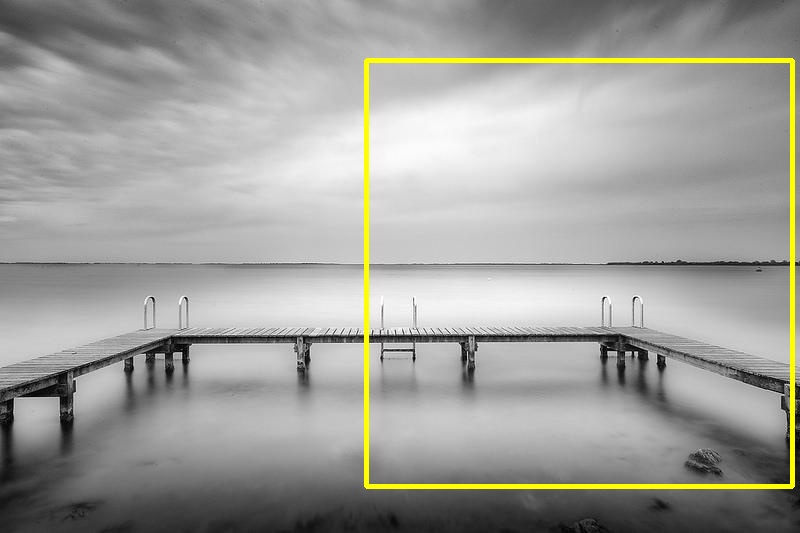}&
\includegraphics[height=1.7cm]{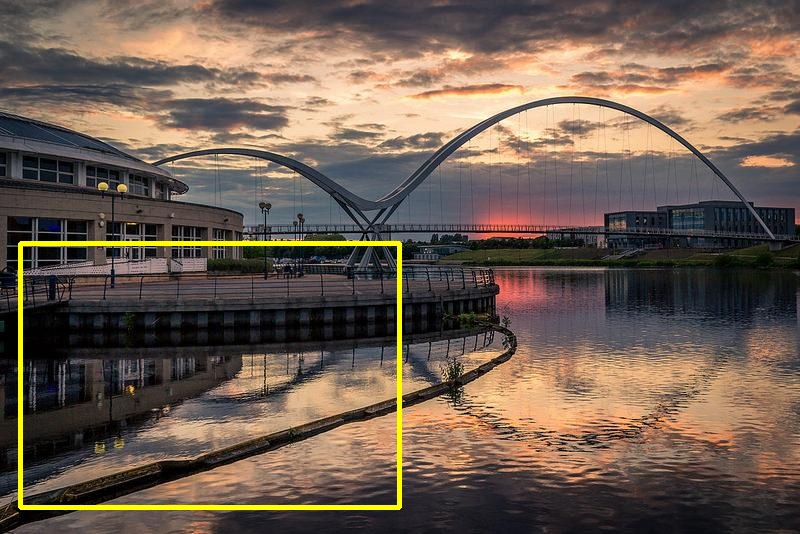}&
\includegraphics[height=1.7cm]{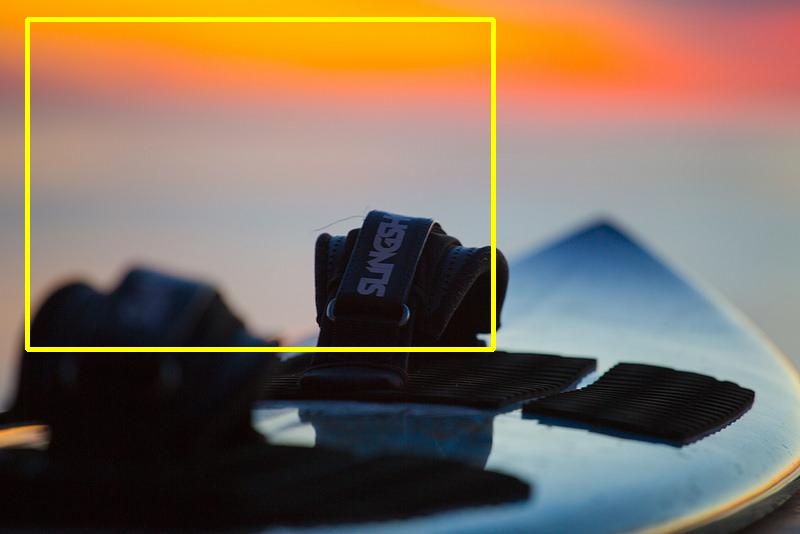}\\
(a) & (b) & (c)
\end{tabular}
\end{center}
\caption{Professional photographs on the web are typically compliant with certain photographic rules. On the other hand, a crop of the image is highly likely to ruin the original composition, \eg, (a) \emph{symmetry}, (b) \emph{rule of thirds}, (c) \emph{object emphasis}. By pairing a professional picture with a corresponding crop, it enables us to leverage human knowledge of photo composition under a learning-to-rank framework.}
\label{fig:crop_examples}
\end{figure}

Rethink about the most basic behavior of photo taking: a photographer repeatedly \emph{moves} the camera\footnote{More specifically, the camera movement may include shift and zoom in/out to properly frame the desired view.} and \emph{judges} if the current view is more visually pleasing than the previous one until the desired view is obtained.
The above observation reveals the essential property of photo composition -- to \emph{successively rank a pair of views with gradually altered contents}.
Unlike most existing methods, which typically try to differentiate the aesthetics of \emph{distinct} images, comparing the aesthetics relationship of visually similar views is relatively easy and less ambiguous.
However, to collect a large amount of ranking samples by human raters for training effective models will inevitably face the aforementioned challenges -- subjectiveness and scalability.

One key observation is that professional photographs are typically compliant with certain photographic rules (see some examples in \figname~\ref{fig:crop_examples}), which are inherently positive examples of good composition.
On the other hand, a crop of professional photographs is highly likely to ruin the original composition.
In other words, a pair of a professional photograph and its corresponding crop is highly likely to possess definite visual preference in terms of aesthetics.
Thanks to the abundant professional photographs on the web, it is thus possible to harvest many unambiguous pairwise aesthetic ranking examples \emph{for free}.
Additionally, the model can naturally learn more photographic rules encoded in more training data without the necessity of explicitly modeling any new hand-crafted features.

Based on the above observations, we formulate the \emph{learning-to-compose} problem as a pairwise view ranking process.
We show that it can be effectively solved by a simple and powerful \emph{view finding network} (VFN), which is trained to honor images of good composition and avoid those of bad composition.
VFN is composed of a widely used object classification network \cite{Krizhevsky:NIPS:2012} optionally augmented with 
a spatial pyramid pooling (SPP) layer \cite{Lazebnik:CVPR:2006,He:ECCV:2014}.
A costless data augmentation method is proposed to collect large-scale ranking samples from the unlimited high-quality images on the web.
Without using any complex hand-crafted features, VFN learns the best photographic practices from examples by \emph{relating} different views in terms of aesthetic ordering.
To evaluate the capability of VFN for view finding, we evaluate its performance on two image cropping databases \cite{Chen:WACV:2017,Yan:CVPR:2013}.
We demonstrate that with simple sliding window search, VFN achieves state-of-the-art performance in cropping accuracy.

To summarize, our main contributions are as follows:
We revisit the extensively studied problem of modeling photo aesthetics and composition and provide new key insights. 
The resulting technical solution is surprisingly simple yet effective.
We show that a large number of automatically generated pairwise ranking constraints can be utilized to effectively train an aesthetics-aware deep ranking network. 
The proposed method significantly outperforms state-of-the-art methods as demonstrated by a quantitative evaluation on two public image cropping datasets.

\section{Previous Work}

Photo composition is an essential factor influencing the aesthetics in photography.
A considerable amount of methods have been developed to assess photo quality \cite{Datta:ECCV:2006,Ke:CVPR:2006,Dhar:CVPR:2011,Luo:ICCV:2011,Nishiyama:CVPR:2011}.
Early works typically exploit ``hand-crafted'' features that mimic certain well-known photographic practices (\eg, rule of thirds, visual balance etc.) and combine them with low-level statistics (\eg, color histogram and wavelet analysis) to accomplish content-based aesthetic analysis.
More recently, generic image descriptors \cite{Marchesotti:ICCV:2011} and deep activation features \cite{Donahue:2013:arXiv} originally targeted at recognition are shown to be generic and outperform rule-based features in aesthetics prediction and style recognition \cite{Karayev:BMVC:2014}.
With the advance of deep learning, recent works \cite{Kang:CVPR:2014,Lu:MM:2014,Lu:ICCV:2015,Mai:CVPR:2016,Kong:ECCV:2016} train end-to-end models without explicitly modeling composition and achieve state-of-the-art performance in the recently released large scale Aesthetics Visual Analysis dataset (AVA) \cite{Murray:CVPR:2012}.

Compared to traditional photo quality assessment methods, which typically exploit photo composition as a high-level cue, some photo \emph{recomposition} techniques attempt to actively enhance image composition by rearranging the visual elements \cite{Bhattacharya:MM:2010}, applying crop-and-retarget operations \cite{Liu:EG:2010} or providing on-site aesthetic feedback \cite{Yao:2012:IJCV} to improve the aesthetics score of the manipulated image.

Photo composition has also been extensively studied in \emph{photo cropping} \cite{Zhang:2013:TIP,Fang:MM:2014,Chen:CVPR:2016} and \emph{view recommendation} \cite{Chang:ICCV:2009,Cheng:MM:2010,Su:TMM:2012} methods.
Generally speaking, these methods aim at the same problem of finding the best view among a number of candidate views within a larger scene and mainly differ in how they differentiate a good view from the bad ones.
Traditionally, \emph{attention-based} approaches exploit visual saliency detection to identify a crop window covering the most visually significant objects \cite{Suh:UIST:2003,Stentiford:ICVS:2007}.
Some hybrid approaches employ a face detector \cite{Zhang:ICME:2005} to locate the region-of-interest or fitting saliency maps to professional photographs \cite{Park:ICIP:2012}.
On the other hand, \emph{aesthetics-based} approaches aim to determine the most visually pleasing candidate window by resorting to photo quality classifiers \cite{Nishiyama:MM:2009}, optimizing composition quality \cite{Fang:MM:2014}, or learning contextual composition rules \cite{Cheng:MM:2010}.
In \cite{Yan:CVPR:2013}, a change-based method is proposed to model the variations before and after cropping so as to discard distracting content and improve the overall composition.
In \cite{Chen:WACV:2017}, the authors first investigate the use of learning-to-rank methods for image cropping.
Unlike our method, they intentionally avoid professional pictures and relied on human raters to rank crops without obvious visual preference, resulting in a moderate-sized database.

To summarize, the main challenges faced by previous methods include 1) the limited applicability of rule-based features, and 2) the difficulty of obtaining composition information for training.
The existing methods or databases build their training data by relying on a few experts \cite{Yan:CVPR:2013,Fang:MM:2014} or crowd-sourcing \cite{Murray:CVPR:2012,Kong:ECCV:2016,Chen:WACV:2017} to annotate and validate the training data, which makes it difficult to scale.
In this work, we tackle these problems with a generic model powered by large-scale training data that is easy to obtain.

\begin{figure}[t]
\begin{center}
\begin{tabular}{c@{\hspace{0.2cm}}c@{\hspace{0.2cm}}c}
\includegraphics[height=2.6cm]{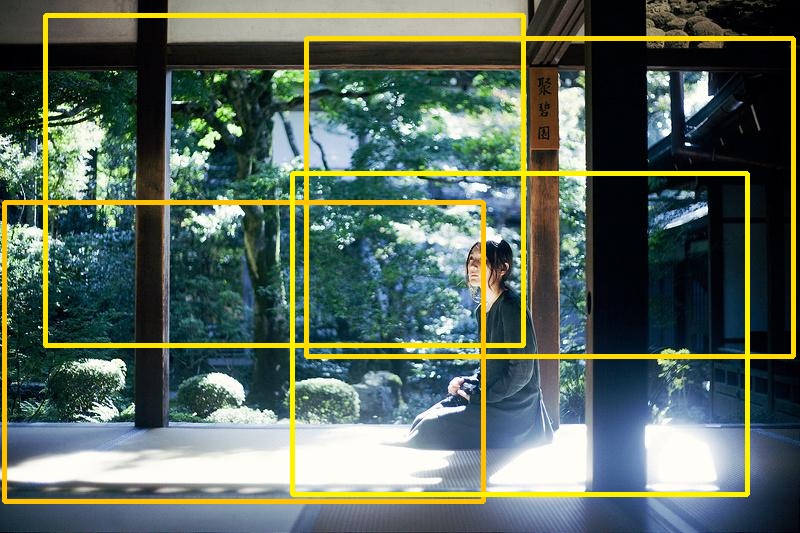}&
\includegraphics[height=2.6cm]{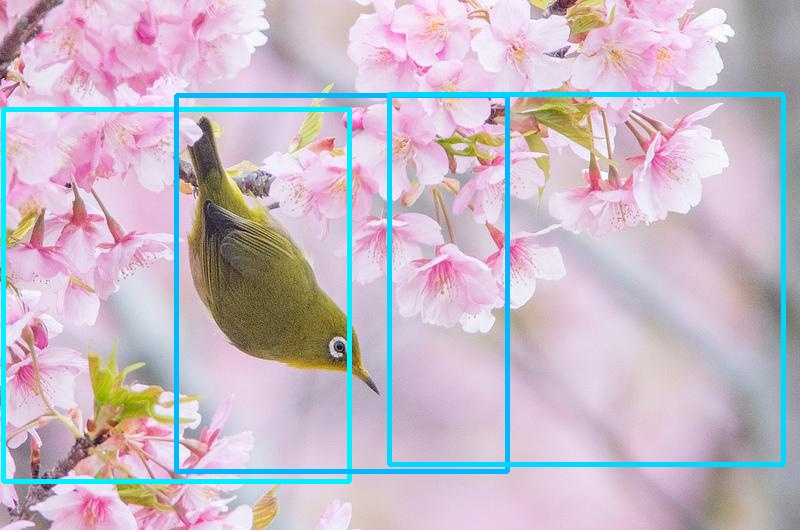}\\
(a) & (b)
\end{tabular}
\end{center}
\caption{Examples of crop generation: (a) border crops, (b) square crops. Best viewed in color. Note that the rectangles  indicate the crops corresponding to a single scale of crops.}
\label{fig:crop_generation}
\end{figure}

\section{Approach}

We model the photo composition or \emph{view finding} process with View Finding Network (VFN).
VFN, which is composed of a CNN augmented with a ranking layer, takes two views as input and predicts the more visually pleasing one in terms of composition.
VFN learns its visual representations (\ie, optimizes the weights of the CNN) by minimizing the misorder of image pairs with known aesthetic preference.
Ideally, by examining extensive examples, VFN learns to compose as human professionals learned their skills.

\subsection{Mining Pairwise Ranking Units}
\label{sec:mining_rank_units}


Every beginner to photography learns by seeing good examples, \ie, professional photographs with perfect composition.
One key observation is that the visual appearance of such golden examples typically achieves a state of \emph{dangerous visual balance}.
It implies that any deviations away from the current view will highly likely degrade the aesthetics -- an inverse process of how the photographer obtained the optimal (current) view.
It is thus possible to costlessly \emph{mine} numerous image pairs with known relative aesthetic ranking.
\figname~\ref{fig:crop_examples} demonstrates several exemplary crops that possess less aesthetics due to violating the photographic heuristics encoded in the original image.

Based on the above observation, we empirically devise the following crop sampling strategies when given a source image $I$:
1) We always form pairs of the original image and a crop because the aesthetic relationship between two random crops is hard to define and thus requires human validation \cite{Chen:WACV:2017}.
2) To enrich the example set required when choosing the best view among different views, we include crops of varying scales and aspect ratios.
3) To best utilize the information in $I$, we aim to maximize the coverage of crops over $I$ while minimizing the overlap between crops.

The resulting crop sampling procedure can be illustrated by \figname~\ref{fig:crop_generation}.
Denote a crop of $I$ as $C$ and $(x, y, w, h)$ indicates its \emph{origin}, \emph{width}, and \emph{height}, respectively.
For each image $I$, we generate a set of \emph{border crops} and \emph{square crops}.
A border crop is created by first placing a uniformly resized window of $I$ at the four corners.
On the other hand, several square crops (we set the number to 3 in our experiments) are created along the long axis of $I$ and evenly spaced.
The parameters of $C$ are then added with a small amount of random perturbation.
Note that the above procedure is by no means the optimal way to generate crops since it is impossible to test all possible configurations.
Nevertheless, different sampling configurations consistently achieve better results than existing methods in our experiments.
Please also refer to the supplementary material for more details of a series of experiments conducted to obtain the crop sampling configurations.

\begin{figure}[t]
\begin{center}
   \includegraphics[width=0.98\linewidth]{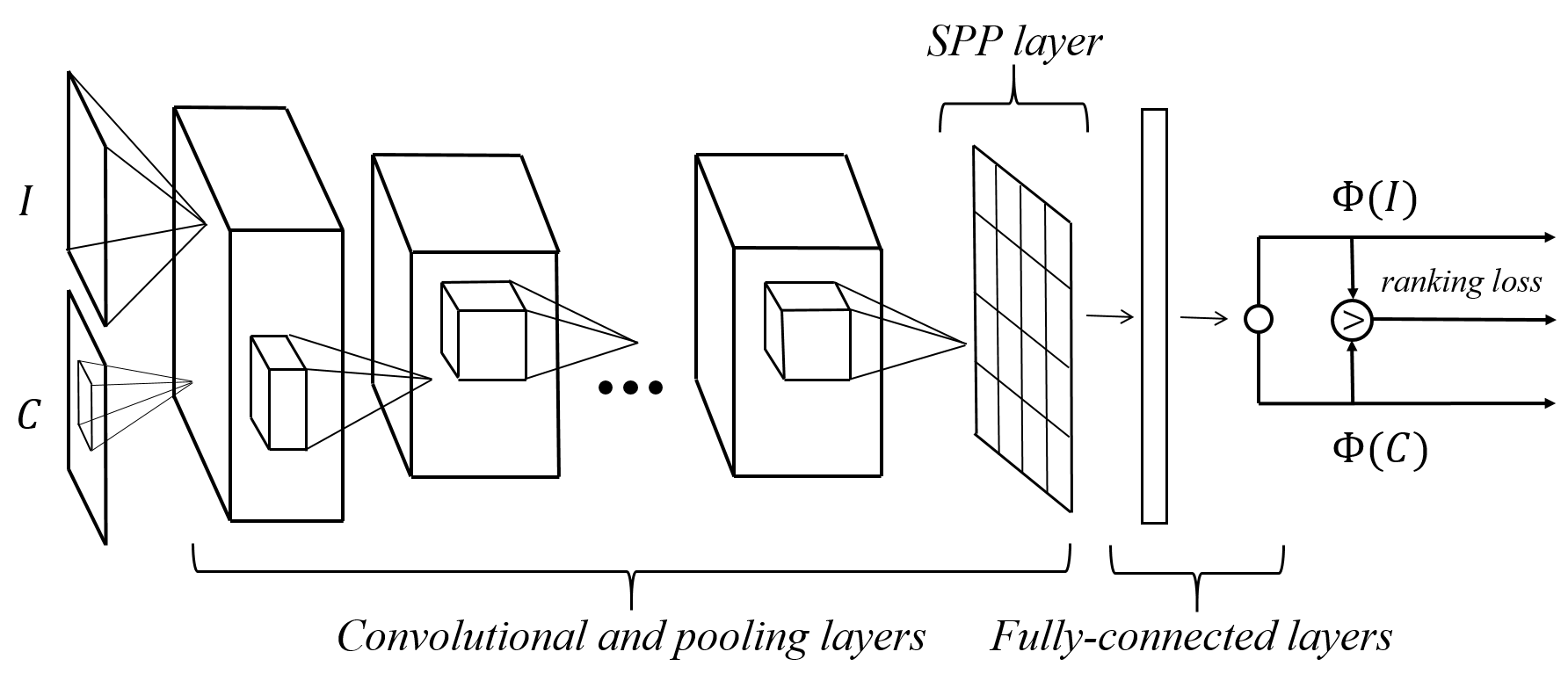}
\end{center}
   \caption{
   Architecture of View Finding Network.
   }
\label{fig:network_architecture}
\end{figure}

\subsection{View Finding Network}

Given an image $I_j$ and its corresponding crops $C^n_j$, the objective of VFN is to learn a mapping function $\Phi(\cdot)$ that relates $I_j$ and $C^n_j$ according to their aesthetic relationship,
\begin{equation}
\Phi(I_j) > \Phi(C^n_j).
\end{equation}
Notice that here we assume that $I_j$ is always higher ranked than $C^n_j$ in terms of aesthetics.
We can thus define the following \emph{hinge loss} for an image pair $(I_j,C^n_j)$:
\begin{equation}
l(I_j,C^n_j)=\max \left\lbrace 0, g+\Phi(C^n_j)-\Phi(I_j)\right\rbrace,
\label{eq:hinge}
\end{equation}
where $g$ is a gap parameter that regularizes the minimal margin between the ranking scores of $I_j$ and $C^n_j$.
We set $g=1$ throughout all experiments.
To learn $\Phi(\cdot)$, we minimize the total loss which sums up $l$ over all training pairs.

Compared to many existing CNN models for aesthetics assessment \cite{Lu:MM:2014,Lu:ICCV:2015,Mai:CVPR:2016,Kong:ECCV:2016}, the architecture of VFN is extremely simple, as illustrated in \figname~\ref{fig:network_architecture}.
The convolutional layers of VFN are adopted from the popular AlexNet \cite{Krizhevsky:NIPS:2012}.
The output of the convolutional layers is then fed into two fully-connected layers followed by a \emph{ranking} layer.
The ranking layer is parameter-free and merely used to evaluate the hinge loss of an image pair.
During training, the model updates its parameters such that $\Phi(\cdot)$ minimizes the total ranking loss in \eqname~(\ref{eq:hinge}).
Once the network is trained, we discard the ranking layer and simply use $\Phi(\cdot)$ to map a given image $I$ to an aesthetic score that differentiates $I$ with other visually similar views.

On top of the last convolutional layer, we \emph{optionally} append a \emph{spatial-pyramid pooling} (SPP) layer \cite{He:ECCV:2014}.
SPP (also known as \emph{spatial pyramid matching} or SPM) \cite{Lazebnik:CVPR:2006} is a widely used method to learn discriminative features by dividing the image with a coarse-to-fine pyramid and aggregating the local features.
It enhances the discrimination power of features by considering the global spatial relations.
Notably, unlike \cite{He:ECCV:2014,Mai:CVPR:2016}, we still use fix-sized input image in VFN (\ie, the input image/patch are first resized to $227\times 227$).
We simply apply the SPP technique to accomplish data aggregation on the convolutional activation features.

Since photo composition is a property affected by both small (\eg, a small object like a flagpole that may destroy the composition) and large structures (\eg, visually significant objects in the scene) in the images,
we thus choose the pooling regions of sizes $3\times 3$, $5\times 5$ and $7\times 7$ with the stride set to one pixel smaller than the pooling size (\eg, 2 for $3\times 3$ pooling regions).
The multi-resolution pooling filters retain composition information at different scales.
In addition, we empirically found that without SPP, the larger feature space causes the model more prone to overfitting.
We apply both \emph{max-pooling} and \emph{average-pooling} in our experiments.

The pooled features are 12,544-dimensional and then fed into the first fully-connected layer.
\texttt{fc1} is followed by a ReLU and has an output dimension of 1,000.
We choose a relatively small feature dimension since the ranking problem is not as complex as object classification.
Besides, as shown in \cite{Chatfield:2014}, convolutional activation features can be compressed without considerable information loss while wide fully connected layers tend to overfit.
\texttt{fc2} has only a single neuron and simply outputs the final ranking scores.

\subsection{Training}

To train our network, stochastic gradient descent algorithm with momentum is employed.
We start from AlexNet \cite{Krizhevsky:NIPS:2012} pre-trained on the ImageNet ILSVRC2012 dataset \cite{ILSVRC15} and the fully connected layers are initialized randomly according to \cite{He:2015}.
Momentum is set to 0.9 and the learning rate starts at 0.01 and is reduced to 0.002 after 10,000 iterations, with each mini-batch comprised of 100 image pairs.
A total of 15,000 iterations is run for training and the validation set is evaluated every 1,000 iterations.
The model with the smallest validation error is selected for testing.
To combat overfitting, the training data is augmented by random horizontal flips as well as slight random perturbations on brightness and contrast.
We implement and train our model with the TensorFlow\footnote{\url{https://www.tensorflow.org/}} framework.

\section{Experimental Results}

\subsection{Training Data}
\label{sec:training_data}

To build the training data, we opt to download pictures shared by professional photographers on the Flickr website\footnote{\url{https://www.flickr.com/}}.
We exploited the Flickr API that returns the ``interesting photos of the day'' and crawled 31,860 images\footnote{We only kept those images with Creative Common license and more than 100 ``favorite'' counts.} during a period of 2,230 consecutive days.
The initial data set is then manually curated to remove non-photographic images (\eg, cartoons, paintings etc.) or images with post-processing affecting the composition (\eg, collage, wide outer frame).
The resulting image pool consists of 21,045 high-quality images and covers the most common categories in photography.
We randomly selected 17,000 images for training and the rest images are used for validation.
As described in Section \ref{sec:mining_rank_units}, we generate 8 border crops and 6 square crops for each image corresponding to two scales $s =$\{0.5, 0.6\}.
Each crop is paired with the corresponding original image and thus there are 294,630 image pairs in total.
The image pair collection is then used to train the VFN.
Note that the above procedure is inexpensive and it is very easy to expand the dataset.

\subsection{Performance Evaluations}

To validate the effectiveness of our model for view finding, we evaluate its \emph{cropping accuracy} on two public image cropping databases, including Flickr Cropping Database (FCDB) \cite{Chen:WACV:2017} and Image Cropping Database (ICDB) \cite{Yan:CVPR:2013}, and compare against several baselines.

\subsubsection{Evaluation Metrics}

We adopt the same evaluation metrics as \cite{Yan:CVPR:2013,Chen:WACV:2017}, \ie, \emph{average intersection-over-union (IoU)} and \emph{average boundary displacement} to measure the \emph{cropping accuracy} of image croppers.
IoU is computed by $area(\hat{C}_i \cap C_i)/area(\hat{C}_i \cup C_i)$,
where $\hat{C}_i$ and $C_i$ denote the ground-truth crop window and the crop window determined by the baseline algorithms for the $i$-th test image, respectively.
Boundary displacement is given by $\sum_{j=1}^4 ||\hat{B}_i^j-B_i^j||/4$,
where $\hat{B}_i^j$ and $B_i^j$ denote the four corresponding edges between $\hat{C}_i$ and $C_i$.
Additionally, we report $\alpha$-\emph{recall}, which is the fraction of best crops that have an overlapping ratio greater than $\alpha$ with the ground truth.
In all of our experiments, we set $\alpha$ to $0.75$.

For the simplicity and fairness of comparison, we follow the sliding window strategy of \cite{Chen:WACV:2017} to evaluate the baselines and VFN.
Similarly, we set the size of search window to each scale among $\left[0.5, 0.6, \dots, 0.9\right]$ of the test images and slide the search window over a 5$\times$5 uniform grid.
The ground truth is also included as a candidate.
The optimal crops determined by individual methods are compared to the ground truth to evaluate their performance.

\subsubsection{Baseline Algorithms}



Following \cite{Yan:CVPR:2013}, we compare with two main categories of traditional image cropping methods, \ie, attention-based and aesthetics-based approaches.
Additionally, we compare with several ranking-based image croppers \cite{Chen:WACV:2017}.
\begin{itemize}
\item Attention-based: For attention-based methods, we choose the best performing method (eDN) reported in \cite{Chen:WACV:2017}, which adopts the saliency detection method described in \cite{Vig:CVPR:2014} and searches for the best crop window that maximizes the difference of saliency score between the crop and the outer region of the image.
\item Aesthetics-based: We choose to fine-tune AlexNet \cite{Krizhevsky:NIPS:2012} for binary aesthetics classification with the AVA dataset \cite{Murray:CVPR:2012} as the baseline of this category and follows the configuration suggested by \cite{Lu:ICCV:2015,Kong:ECCV:2016}.
We simply utilize the softmax confidence score to choose the best view.
The methods of \cite{Yan:CVPR:2013,Mai:CVPR:2016} also fall into this category.
We compare with these methods by the accuracy reported in the original paper \cite{Yan:CVPR:2013} or use the pre-trained model to evaluate its performance in both datasets \cite{Mai:CVPR:2016}.
\item Ranking-based: We adopt two variants of RankSVM-based image croppers using deep activation features \cite{Donahue:2013:arXiv} and trained on the AVA and FCDB datasets \cite{Chen:WACV:2017}, which differ in their data characteristics.
AVA characterizes the aesthetics preference between \emph{distinct} images while FCDB provides the ranking order between crop pairs in the \emph{same} images.
Additionally, we compare with the recent work of aesthetics ranking network \cite{Kong:ECCV:2016}.
We use the pre-trained model released by the authors and utilize the ranking scores of the sliding windows to determine the best crop.
\end{itemize}

\begin{table}[t]
\begin{center}
\begin{tabular}{|c||c|c|c|}
\hline
Method                                  & IoU     &  Disp.   &  $\alpha$-recall  \\
\hline
eDN \cite{Vig:CVPR:2014}                & 0.4929  &  0.1356  &  12.68    \\
AlexNet\_finetune                       & 0.5543  &  0.1209  &  16.092   \\
MNA-CNN \cite{Mai:CVPR:2016}            & 0.5042  &  0.1361  &  0.0747   \\
RankSVM+AVA \cite{Chen:WACV:2017}       & 0.5270  &  0.1277  &  12.6437  \\
RankSVM+FCDB \cite{Chen:WACV:2017}      & 0.602   &  0.1057  &  18.1034  \\
AesRankNet \cite{Kong:ECCV:2016}        & 0.4843  &  0.1401  &  0.0804   \\ \hline
VFN                                     & \textbf{0.6842}  &  \textbf{0.0843}  &  \textbf{35.0575}  \\
VFN+AVA (SPP-Max)                       & 0.544   &  0.124   &  12.93    \\
VFN (SPP-Avg)                           & 0.6783  &  0.0859  &  35.0575  \\
VFN (SPP-Max)                           & 0.6744  &  0.0872  &  33.9080  \\
\hline
\end{tabular}
\end{center}
\caption{Performance comparison on FCDB \cite{Chen:WACV:2017}.
The best results are highlighted in bold.
}
\label{tab:cropping_performance_FCDB}
\end{table}

\begin{table*}[t]
\begin{center}
\begin{tabular}{@{}|c||c|c|c||c|c|c||c|c|c|@{}}
\hline
\multirow{2}{*}{Method} & \multicolumn{3}{c||}{Annotation Set \#1} & \multicolumn{3}{c||}{Annotation Set \#2} & \multicolumn{3}{c|}{Annotation Set \#3} \\ \cline{2-10}
                                       &  IoU      &  Disp.   &  $\alpha$-recall  &  IoU     &  Disp.   &  $\alpha$-recall  &  IoU     &  Disp.   &  $\alpha$-recall  \\ \hline
  eDN \cite{Vig:CVPR:2014}             &  0.5535   &  0.1273  &  27.3684          &  0.5128  &  0.1419  &  20.1053          &  0.5257  &  0.1358  &  22.4211          \\
  AlexNet\_finetune                    &  0.5687   &  0.1246  &  23.0526          &  0.5536  &  0.1296  &  22.7368          &  0.5544  &  0.1288  &  20.6316          \\
  MNA-CNN \cite{Mai:CVPR:2016}         &  0.4693   &  0.1555  &  0.0716           &  0.4553  &  0.1615  &  0.0642           &          0.4610  &  0.1590  &  0.0684           \\
  RankSVM+AVA \cite{Chen:WACV:2017}    &  0.5801   &  0.1174  &  18.7368          &  0.5678  &  0.1225  &  18.6316          &  0.5665  &  0.1226  &  18.9474          \\
  RankSVM+FCDB \cite{Chen:WACV:2017}   &  0.6683   &  0.0907  &  33.4737          &  0.6618  &  0.0932  &  32.1053          &  0.6483  &  0.0973  &  31.2632          \\
  AesRankNet \cite{Kong:ECCV:2016}     &  0.4484   &  0.1631  &  0.0863           &  0.4372  &  0.168   &  0.0747           &  0.4408  &  0.1655  &  0.0863           \\
  LearnChange \cite{Yan:CVPR:2013}     &  0.7487   &  0.0667  &   --              &  0.7288  &  0.072   &  --               &  0.7322  &  0.0719  &  --      \\ \hline
  VFN                                  &  0.7720   &  0.0623  &  58.8421          &  0.7638  &  0.0654  &  56.4211          &  0.7487  &  0.0692  &  53.7895          \\
  VFN+AVA (SPP-Max)                    &  0.5273   &  0.1387  &  18.21            &  0.5268  &  0.14    &  19.0526          &  0.5261  &  0.1389  &  18               \\
  VFN (SPP-Avg)                         &  0.7837   &  0.0588  &  \textbf{61.5789}          &  0.7729  &  0.0627  &  58.1053          &  0.7514  &  0.0681  &  54.1053          \\
  VFN (SPP-Max)                         &  \textbf{0.7847}   &  \textbf{0.0581}  &  59.7895          &  \textbf{0.7763}  &  \textbf{0.0614}  &  \textbf{58.1053}          &  \textbf{0.7602}  &  \textbf{0.0653}  &  \textbf{54.8421}          \\
\hline
\end{tabular}
\end{center}
\caption{Performance evaluation on ICDB \cite{Yan:CVPR:2013}. The best results are highlighted in bold.}
\label{tab:cropping_performance_ICDB}
\end{table*}

\subsubsection{Performance Evaluation}
\label{sec:performance_evaluation}

We evaluate cropping accuracy of VFN and several baselines on FCDB and ICDB, which differ in data characteristics and annotation procedure.
The test set of FCDB contains 348 images.
Each image was labeled by a photography hobbyist and then validated by 7 workers on Amazon Mechanical Turk.
On the other hand, ICDB includes 950 images, each annotated by 3 experts.
The images of ICDB are typically of \emph{iconic} views and thus more object-centric.
Compared to ICDB, FCDB is considered to be more challenging for image cropping methods because the annotations reflect the tastes of various photographers and the images contain more contextual information.

\tabname~\ref{tab:cropping_performance_FCDB} and \ref{tab:cropping_performance_ICDB} summarize the benchmark results.
Generally, the performance of each category is consistent with \cite{Chen:WACV:2017}.
The attention-based method (eDN) performs poorly due to the lack of aesthetic consideration and
aesthetics-based methods based on a photo quality classifier (AlexNet\_finetune) achieves only moderate performance.
Surprisingly, the aesthetics ranking network \cite{Kong:ECCV:2016} and MNA-CNN \cite{Mai:CVPR:2016} methods also do not perform well in the benchmark.
This is most possibly because these networks are trained to predict the aesthetic rating of distinct images, which does not reflect the relations between different views with large overlaps.
We validate this by training VFN with the traditional dataset \cite{Murray:CVPR:2012} and will discuss the results soon.
Additionally, some image attributes (\eg, color) assessed by the model may not be very discriminating for similar views.

All variants of VFN trained by our data sampling technique significantly outperform the other baselines.
The best performing baseline method is the change-based algorithm \cite{Yan:CVPR:2013}, which achieves very good results in their dataset (ICDB).
Notably, the model of \cite{Yan:CVPR:2013} is trained on the first annotation set and evaluated on the same images of all annotation sets.
On the other hand, the images of ICDB are totally unseen to our models.
In addition, a crop selection procedure which selects an initial set of good candidate windows is incorporated in \cite{Yan:CVPR:2013}, while VFN is evaluated by a fixed set of sliding windows.

\begin{figure}[t]
\begin{center}
   \includegraphics[width=0.98\linewidth]{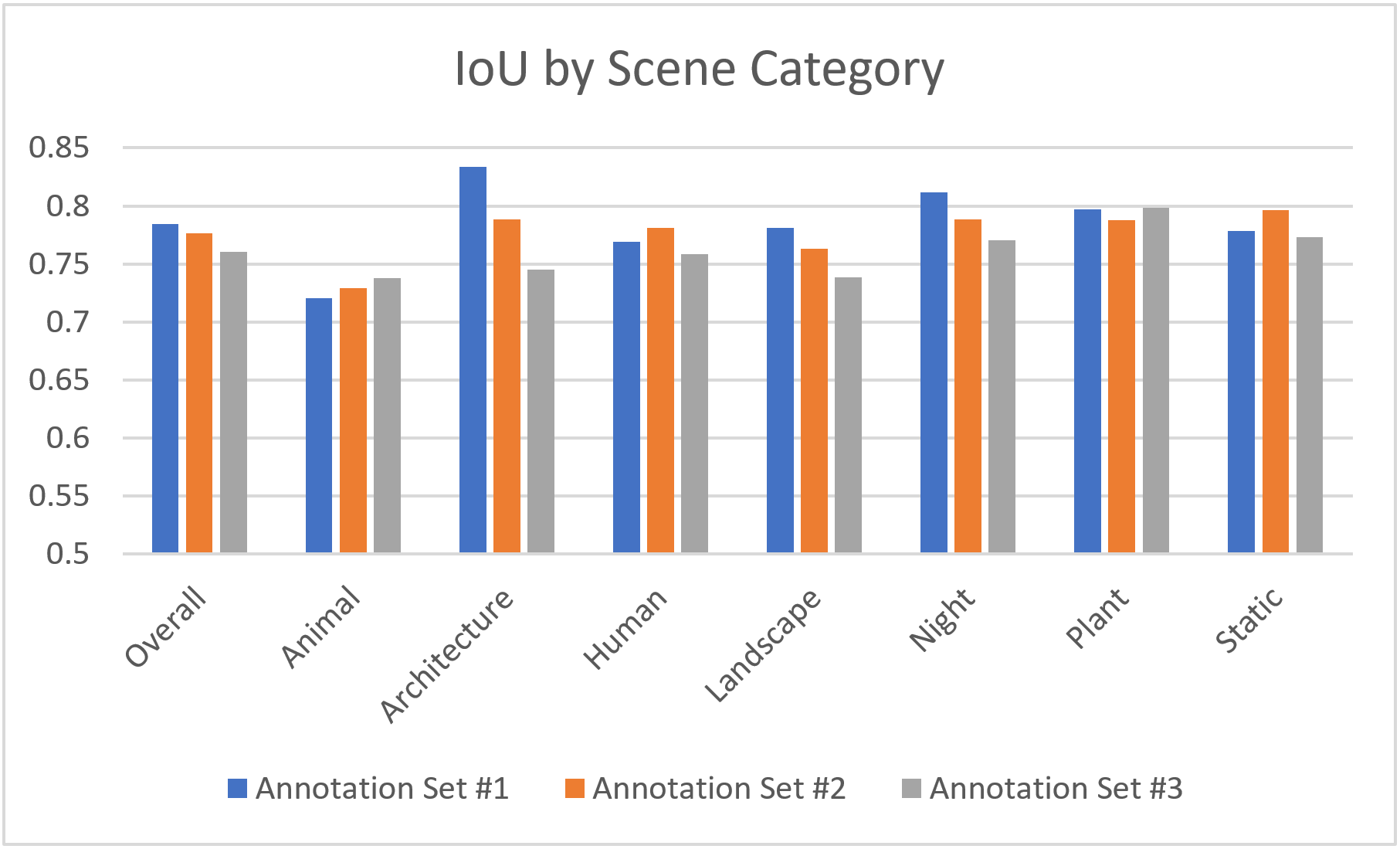}
\end{center}
   \caption{
   Performance of VFN (SPP-Max) on ICDB by category.
   }
\label{fig:category_performance}
\end{figure}

We conduct a more fine-grained performance analysis of VFN by the scene category annotations provided by ICDB, which further divide the dataset into seven categories: \emph{animal}, \emph{architecture}, \emph{human}, \emph{landscape}, \emph{night}, \emph{plant} and \emph{static}.
\figname~\ref{fig:category_performance} illustrates the average IoU scores of VFN (SPP-Max) over various annotation sets and scene categories.
We consider VFN as a generic aesthetics model since it shows no \emph{bias} to specific annotation sets or categories.
It performs generally well across subjects despite image cropping's subjective nature.
Since no category-specific features are exploited, VFN generalizes across categories as well.
Nevertheless, we still can see some insufficiency of VFN, \eg, consistently lower accuracy in \emph{animal} and higher performance variance in \emph{architecture}.
Considering the data-driven nature of VFN, this phenomenon can be most possibly accounted for insufficient or unbalanced distribution of training images among various categories.

Some more interesting observations can be made from the benchmarking results, as discussed below.

\emph{View finding is intrinsically a problem of ranking pairwise views in the same context.}
Intuitively, image rankers trained on aesthetics relations derived from distinct images, such as RankSVM+AVA and \cite{Kong:ECCV:2016}, do not necessarily perform well in ranking visually similar views.
To further validate such an assumption, we additionally trained a VFN with ranking units purely sampled from AVA.
To mitigate the ambiguity of ranking relationship between images, we choose the 30,000 highest and lowest ranked images from AVA and randomly select a pair of images from each pool to train VFN.
The resulting model (VFN+AVA) can be regarded as the counterpart of RankSVM+AVA.
As shown in \tabname~\ref{tab:cropping_performance_FCDB} and \ref{tab:cropping_performance_ICDB}, the performance of VFN+AVA drastically degrades compared to other variants of VFN.
It also confirms that our data sampling technique contributes to the most significant leap in performance.

\emph{Performance gain due to top level pooling is dependent on the characteristics of test data.}
VFN achieves the best results in ICDB and FCDB with and without SPP, respectively.
Recall that the images in ICDB are largely object-centric with iconic views.
Since pooled features are typically equipped with certain \emph{invariance} (\eg, translation), it is thus beneficial to discriminate scenes with significant objects. 
On the other hand, since the images in FCDB possess richer contextual information, 
the greater feature space without pooling is thus more capable of capturing more subtle variations in photo composition,
resulting in higher performance.

\subsection{Applications}

\paragraph{Automatic image cropping}
The ability of VFN makes it very suitable to facilitate the process of identifying unattractive regions in an image to be cut away so as to improve its visual quality.
As demonstrated by the quantitative evaluation in \secname~\ref{sec:performance_evaluation}, VFN achieves state-of-the-art performance in two image cropping datasets.
\figname~\ref{fig:baseline_comparison} illustrates several examples of applying VFN to crop images from FCDB and compares the results with several baselines.
One can see that VFN successfully selects more visually pleasing crop windows compared to other baseline algorithms.
Some of the results by VFN are arguably no worse than the ground truth (\eg, the 2nd and 3rd row in \figname~\ref{fig:baseline_comparison}).
Currently, only sliding windows with the same aspect ratio as the original image are used for evaluation, which limits VFN's ability to identify other possible good compositions, as the ground truth shown in the 1st row of \figname~\ref{fig:baseline_comparison}.
Nevertheless, VFN selects a preferable view with rule-of-thirds composition in this example when compared with other baselines.
However, a crop selection procedure that adaptively determines the parameters of crop windows is still desirable for VFN to maximize its performance.

\begin{figure}[t]
\begin{center}
\begin{tabular}{c@{\hspace{0.2cm}}c@{\hspace{0.2cm}}c}
\includegraphics[height=2.4cm]{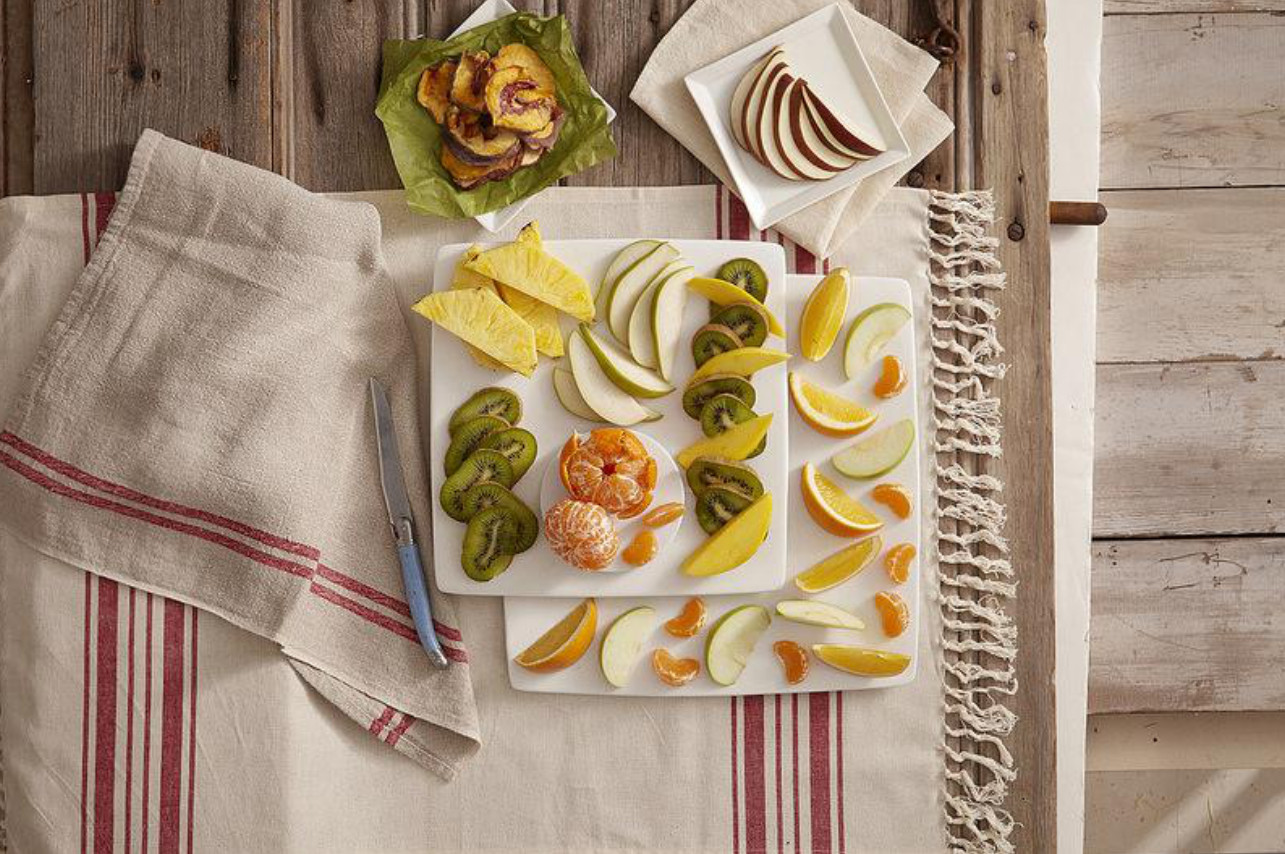}&
\includegraphics[height=2.4cm]{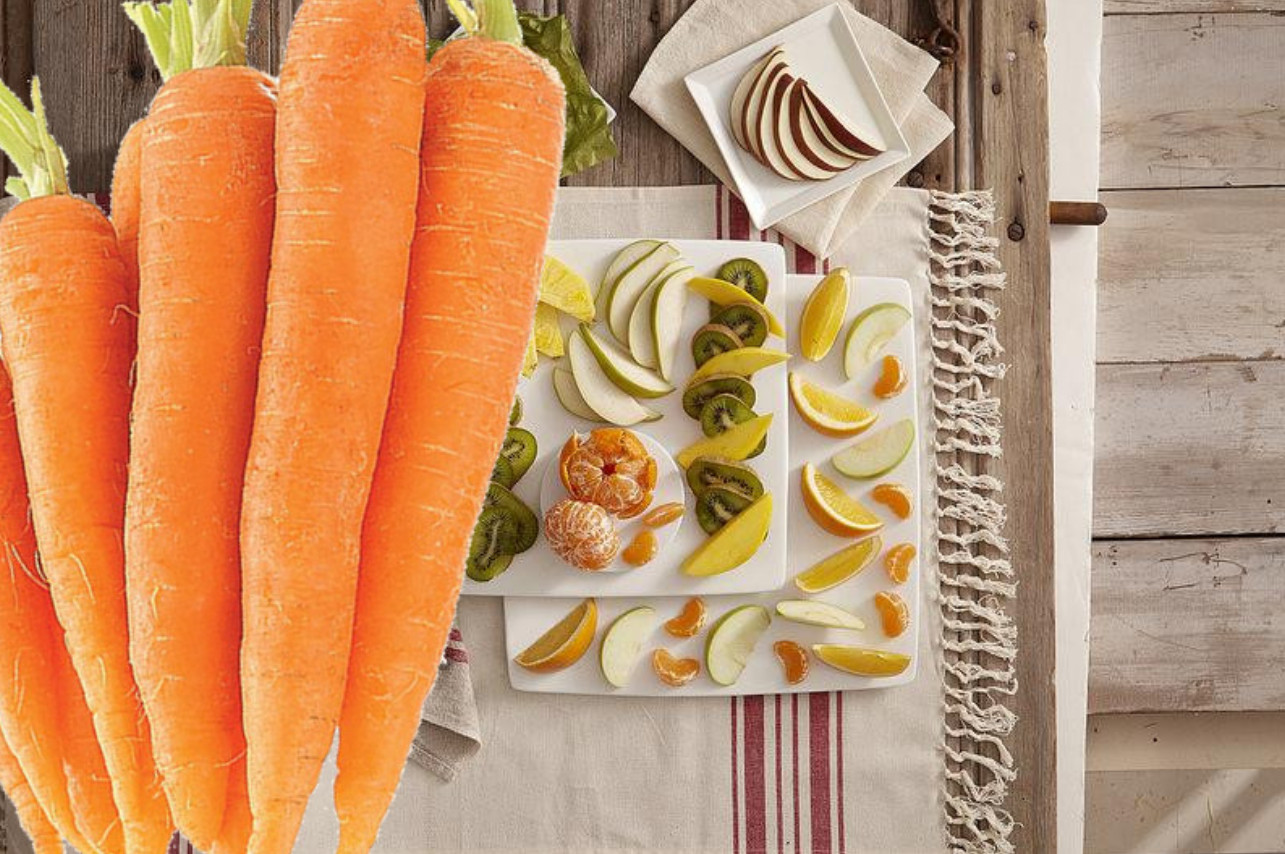}\\
(a) & (b)\\
\includegraphics[height=2.4cm]{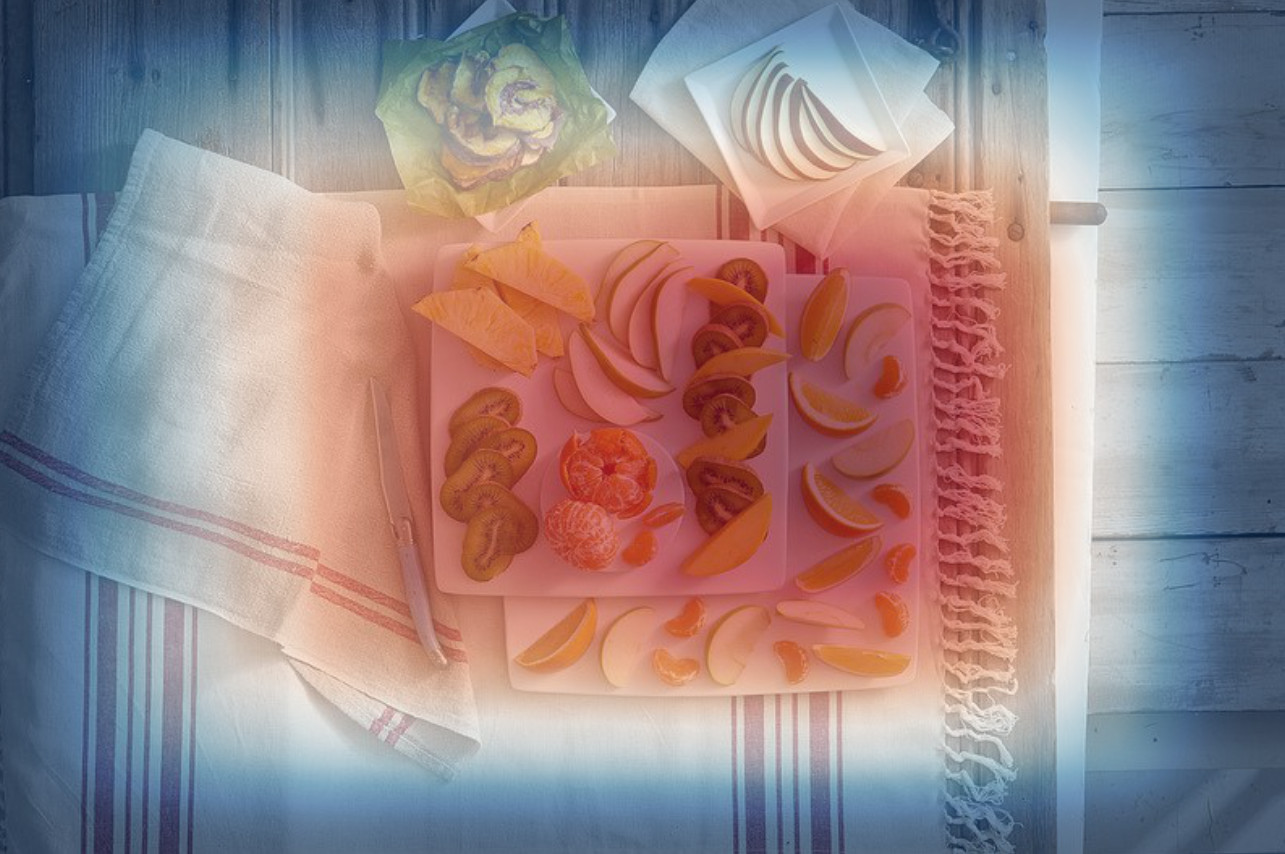}&
\includegraphics[height=2.4cm]{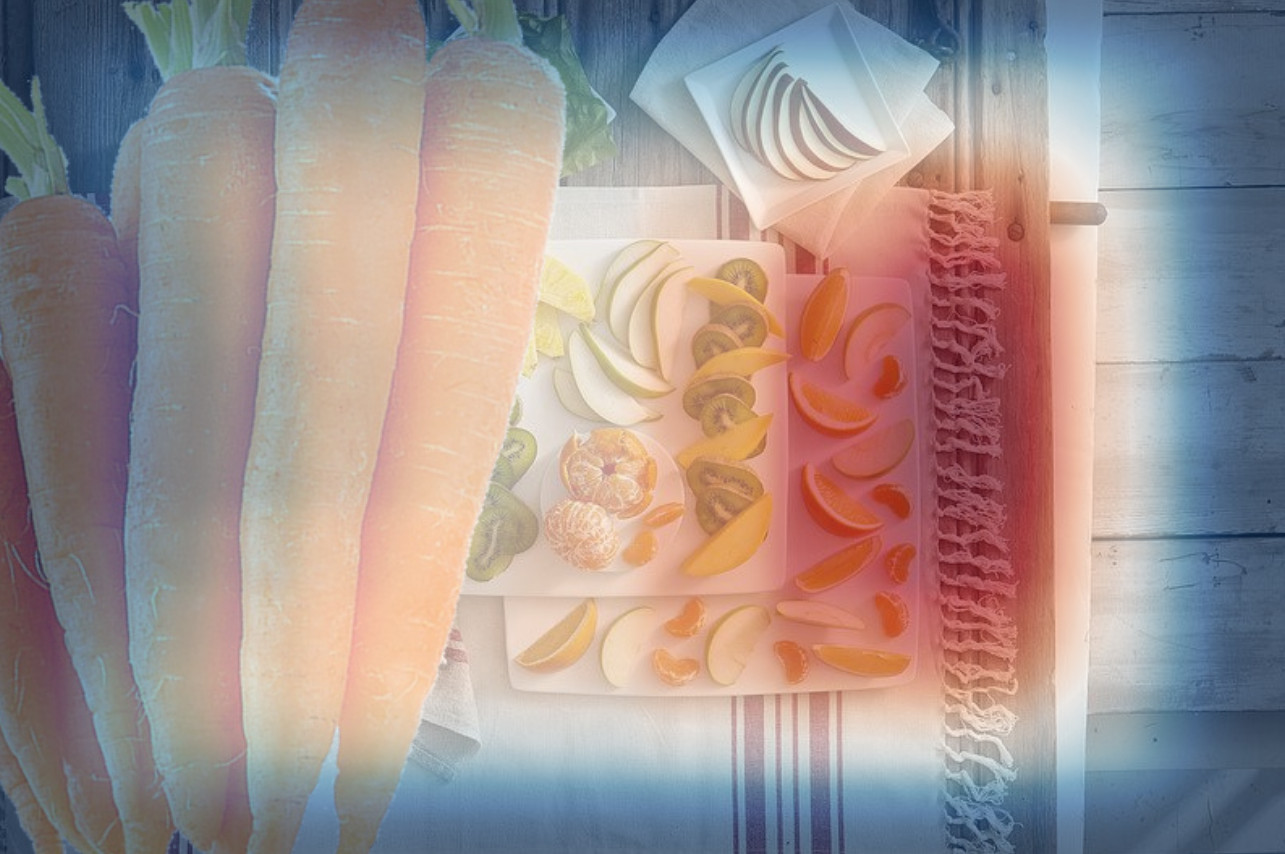}\\
(c) & (d)
\end{tabular}
\end{center}
\caption{VFN is aesthetics-aware and capable of differentiating good/bad views in terms of photo composition. Given an source image (a) and a corrupted image (b), VFN produces higher response to the visually pleasing regions, as demonstrated in the corresponding heatmaps (c)(d).}
\label{fig:heatmap}
\end{figure}

\begin{figure*}[t]
\begin{center}
\includegraphics[height=3.4cm]{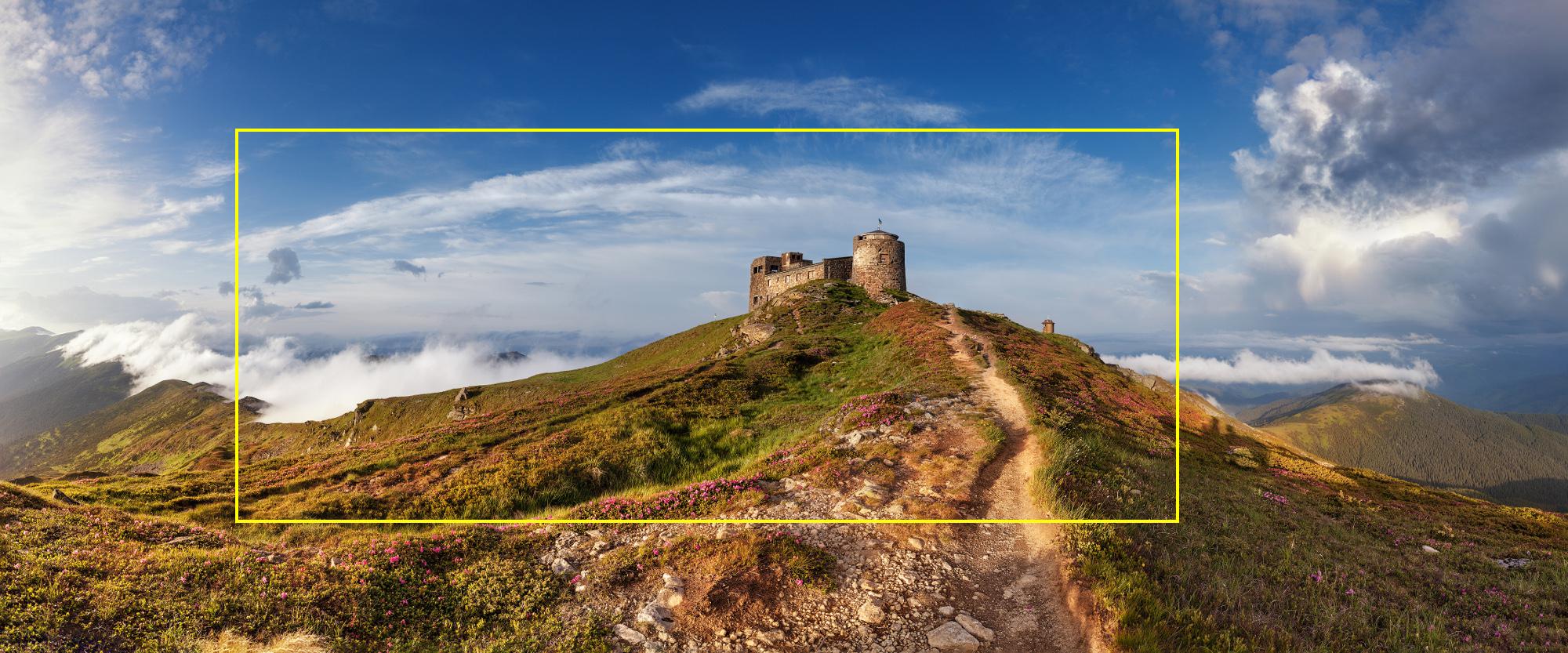}~
\includegraphics[height=3.4cm]{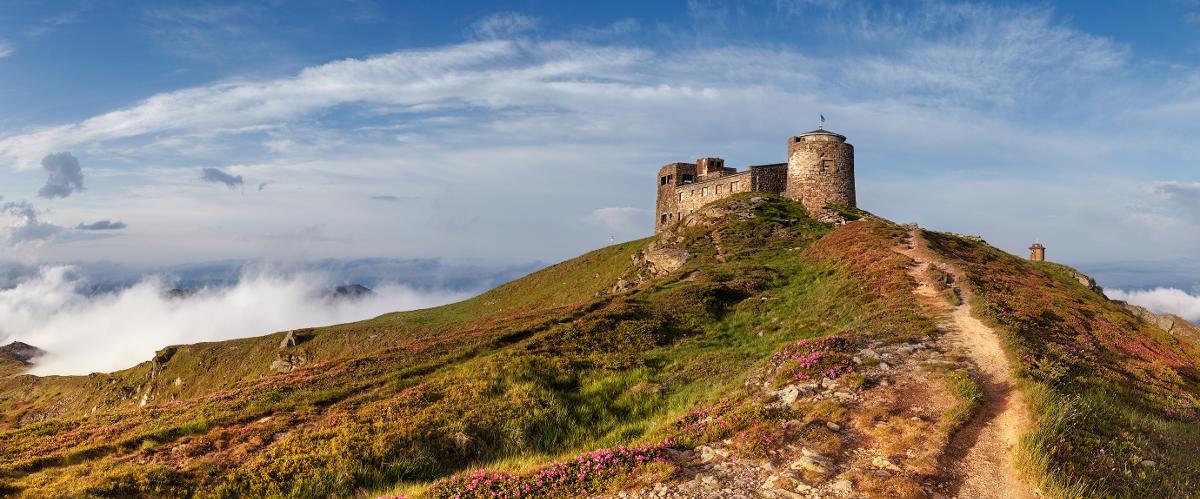}\\
(a)\\
\includegraphics[height=2.8cm]{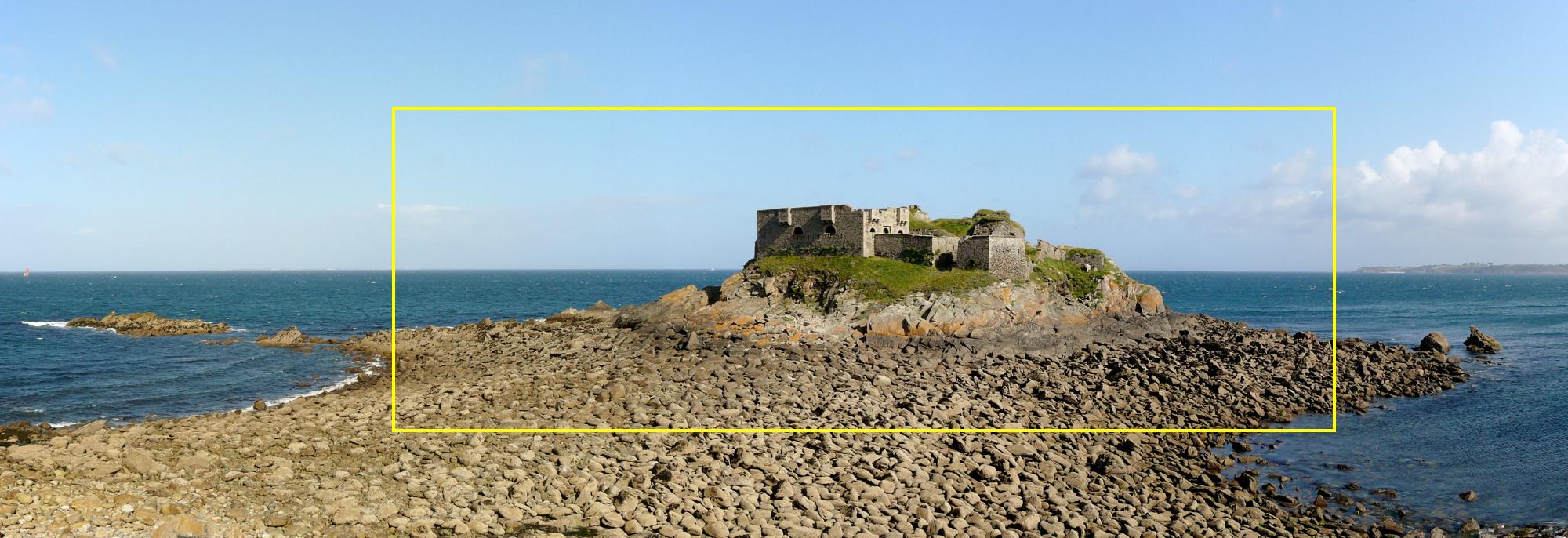}~
\includegraphics[height=2.8cm]{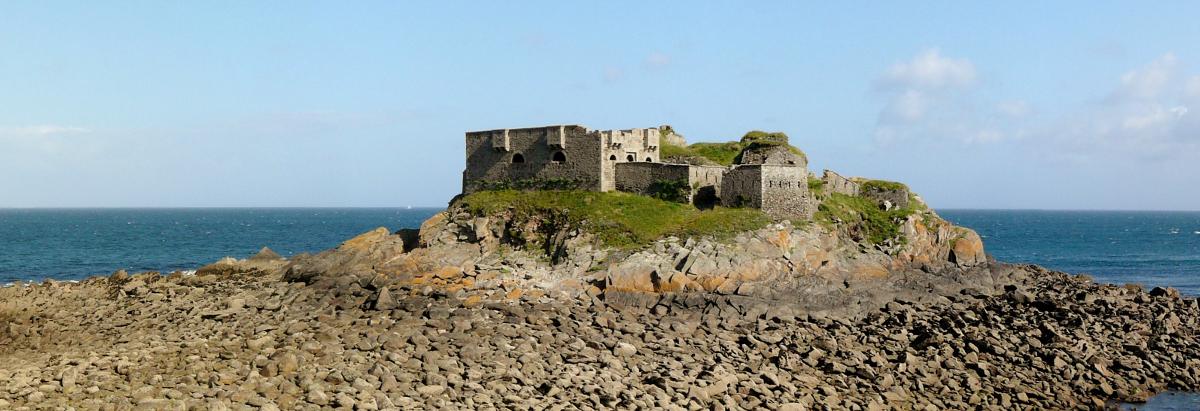}\\
(b)\\
\includegraphics[height=3.27cm]{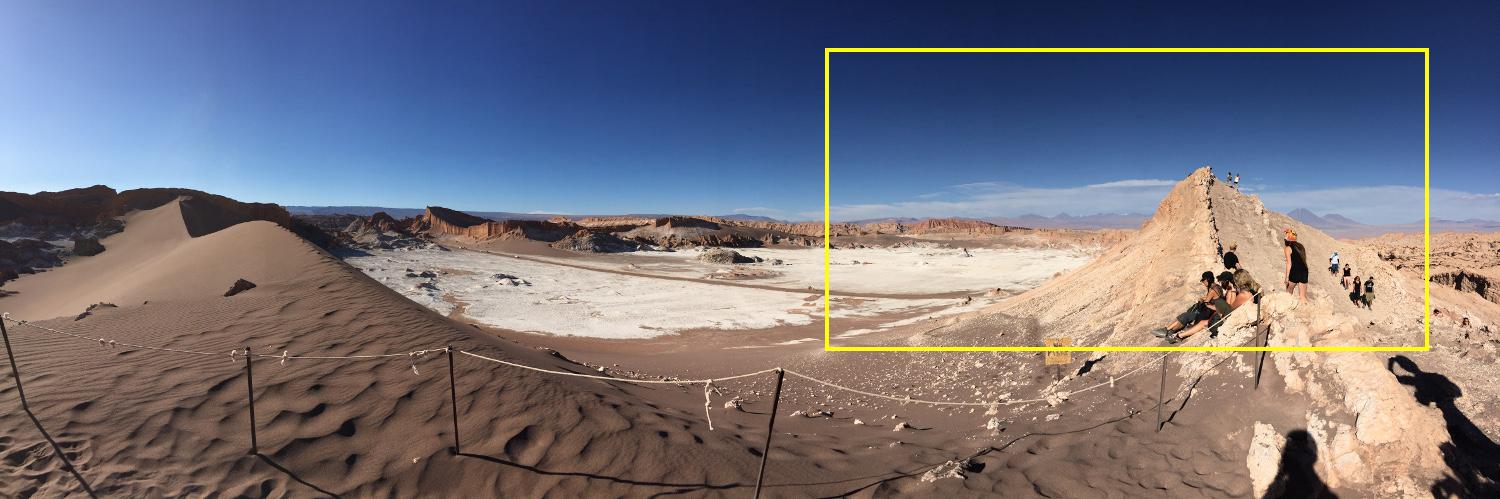}~
\includegraphics[height=3.27cm]{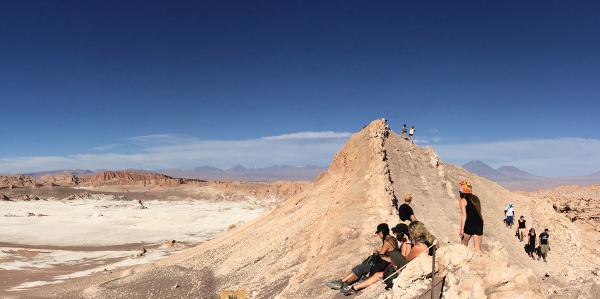}\\
(c)\\
\includegraphics[height=3.65cm]{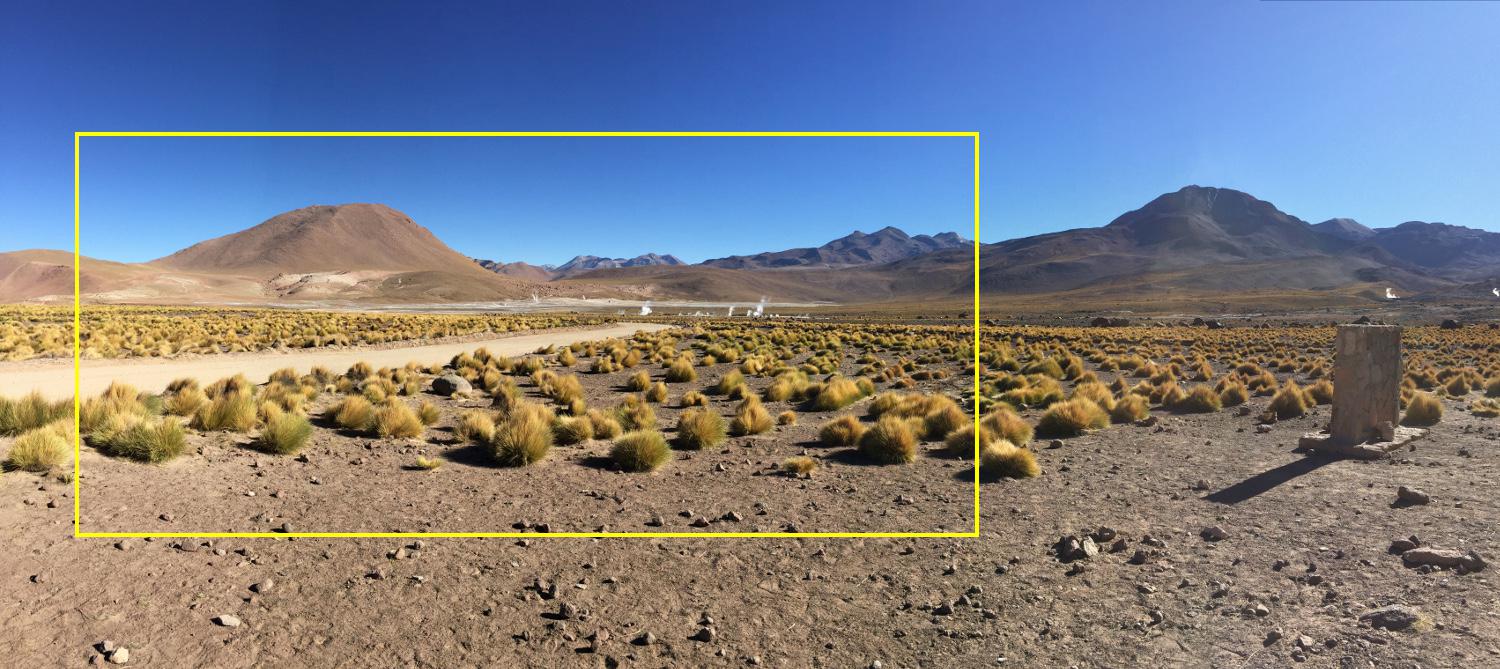}~
\includegraphics[height=3.65cm]{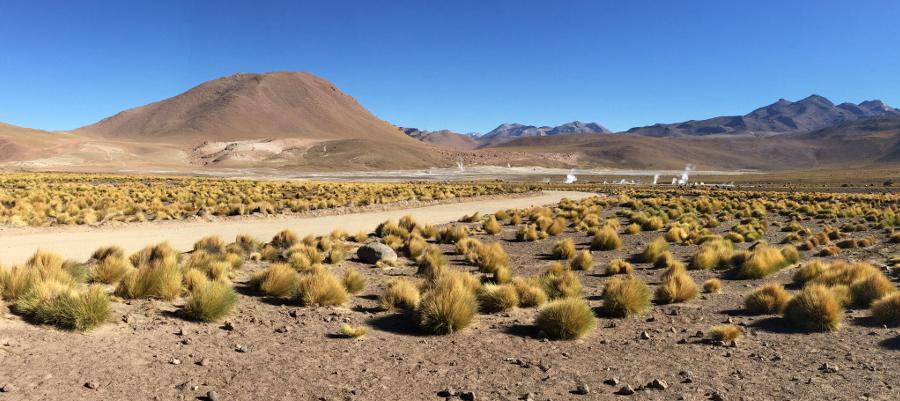}\\
(d)
\end{center}
\caption{An example of applying VFN to a panorama image. The yellow rectangles in the left column images indicate the crop with the maximum score among 2112 uniformly sampled candidate crops of different sizes and aspect ratios. The resulting crop is shown in right column.
Best viewed in color.
}
\label{fig:panorama}
\end{figure*}

\paragraph{View recommendation}
VFN is aesthetics-aware and very sensitive to the variation of image composition.
\figname~\ref{fig:heatmap} demonstrates an example of applying VFN to an image and its artificially ``corrupted'' version.
We generate a heatmap by evaluating sliding windows and smoothing the ranking scores corresponding to the raw pixels.
As one can see, the altered image composition causes VFN to shift its attention to the untouched region.
Due to its aesthetics-awareness, VFN is very suitable to be applied for view suggestion in panoramic scenes or even 360 video, as demonstrated in \figname~\ref{fig:panorama}.
In this example, VFN identifies a visually attractive view while ignoring large unimportant areas in the scene.
Unlike \cite{Chang:ICCV:2009}, which requires a \emph{query} or \emph{template} image to locate similar views in the panoramic image, our model is able to suggest a good view based on a much larger \emph{database} (\ie, the training images).


\subsection{Discussion}

\begin{figure*}[ht!]
\begin{center}
\begin{tabular}{c@{\hspace{0.2cm}}c@{\hspace{0.2cm}}c@{\hspace{0.2cm}}c@{\hspace{0.2cm}}c@{\hspace{0.2cm}}}

\includegraphics[width=3.2cm]{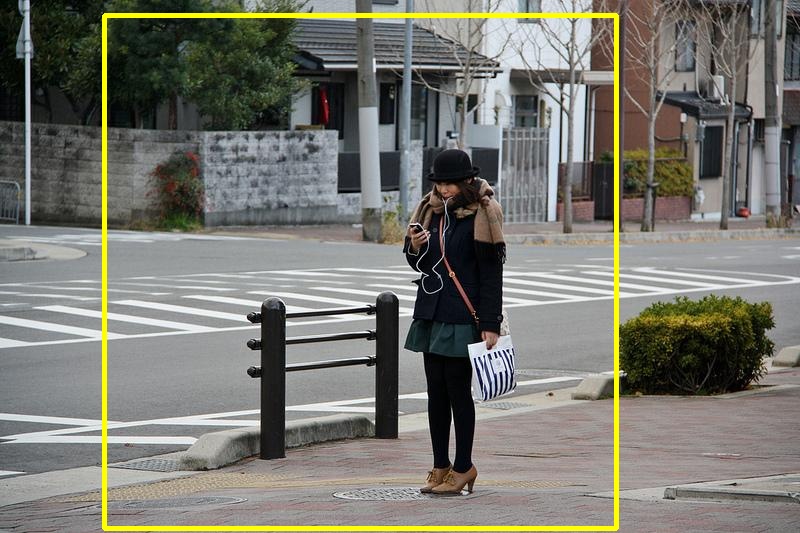}&
\includegraphics[width=3.2cm]{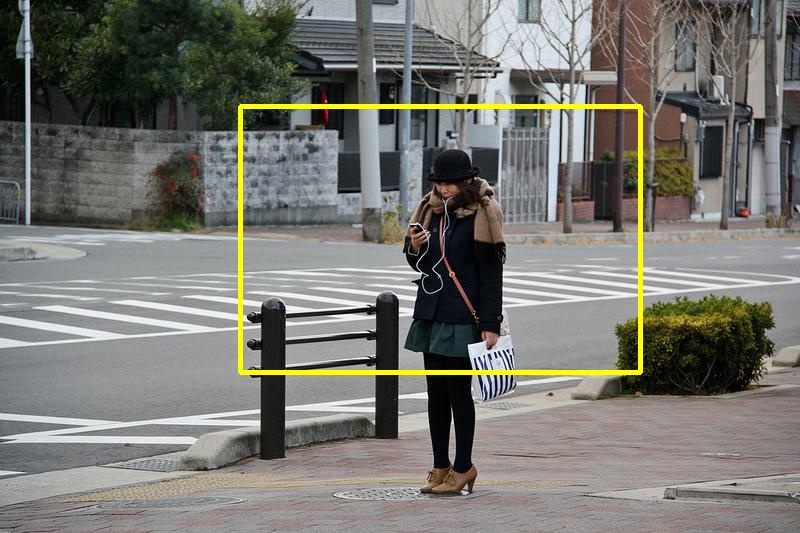}&
\includegraphics[width=3.2cm]{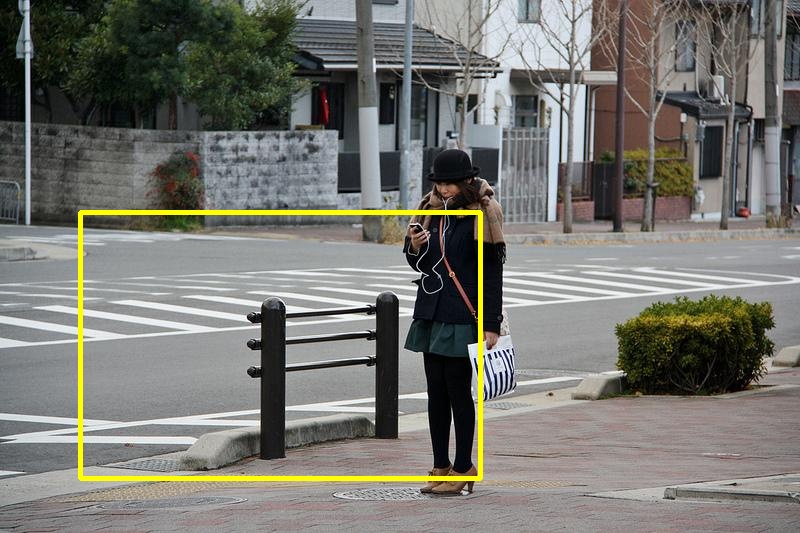}&
\includegraphics[width=3.2cm]{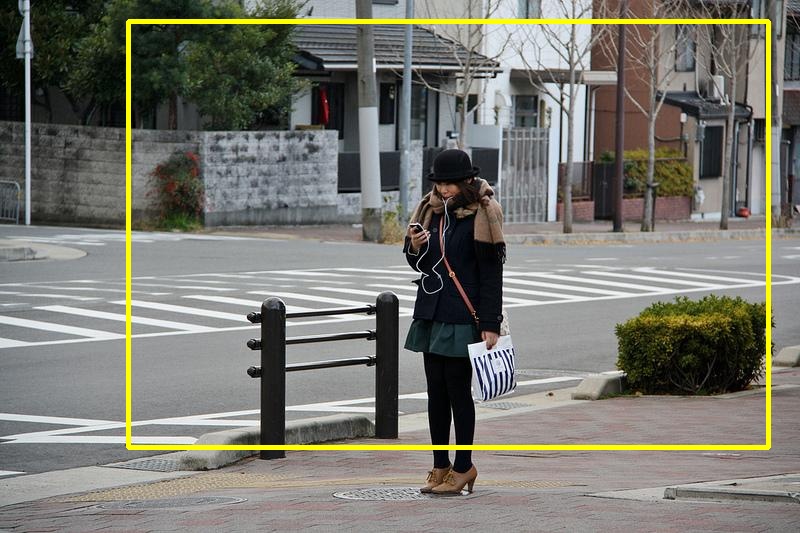}&
\includegraphics[width=3.2cm]{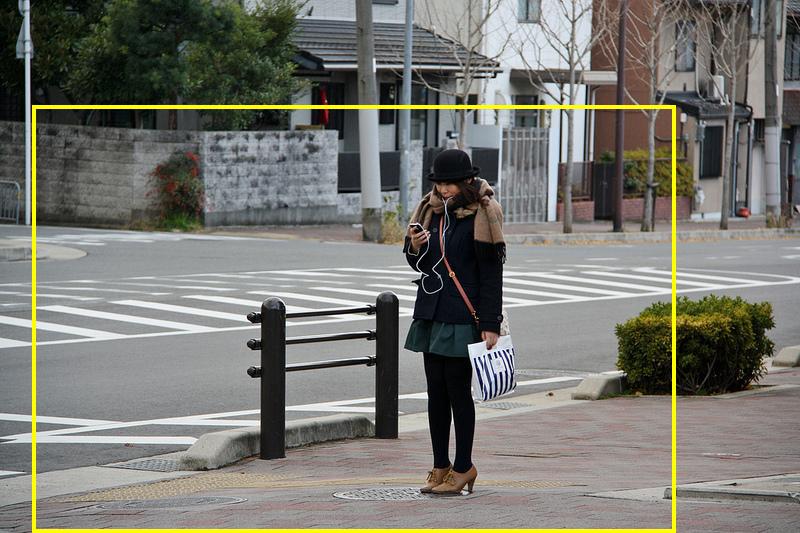}\\

\includegraphics[width=3.2cm]{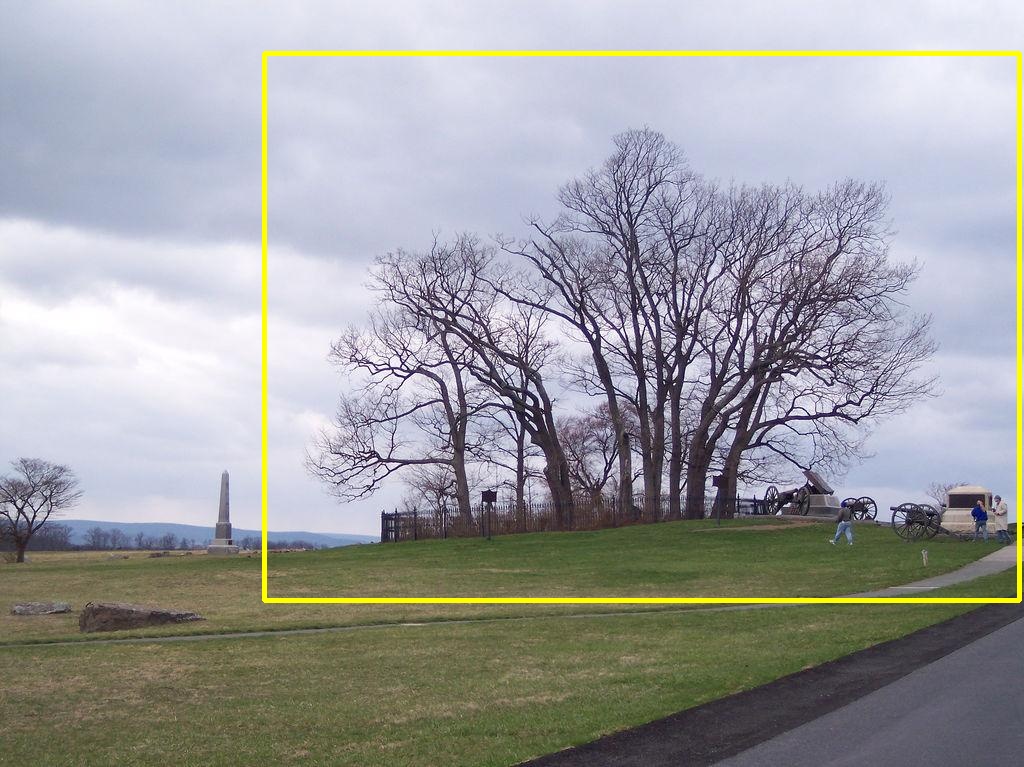}&
\includegraphics[width=3.2cm]{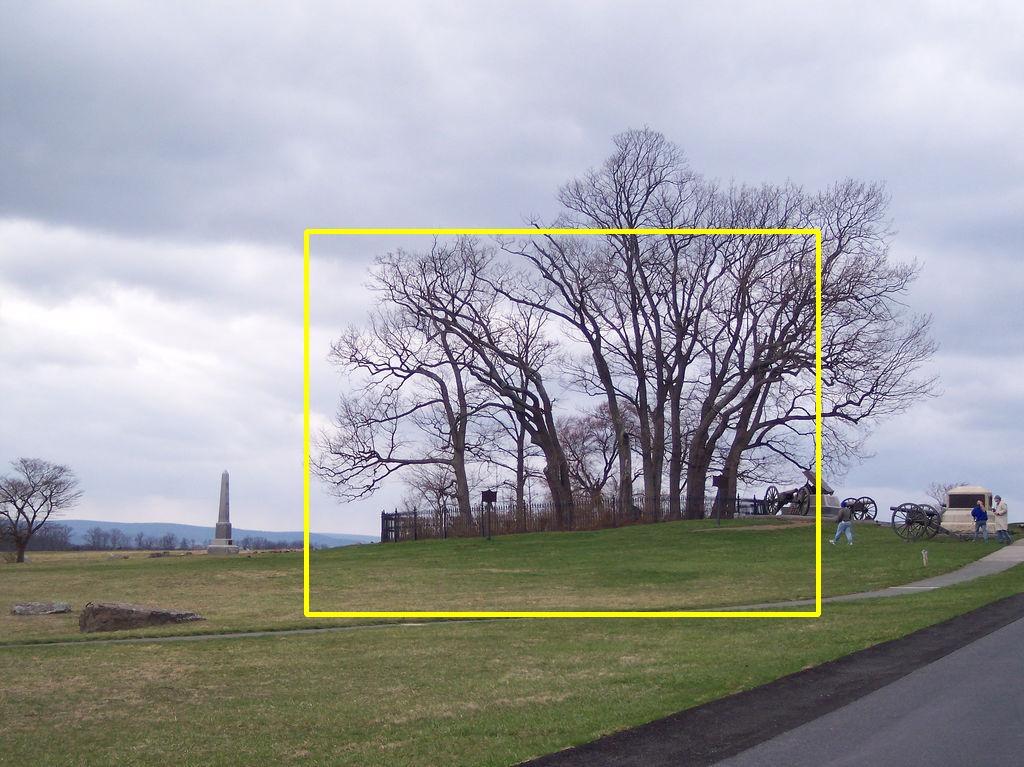}&
\includegraphics[width=3.2cm]{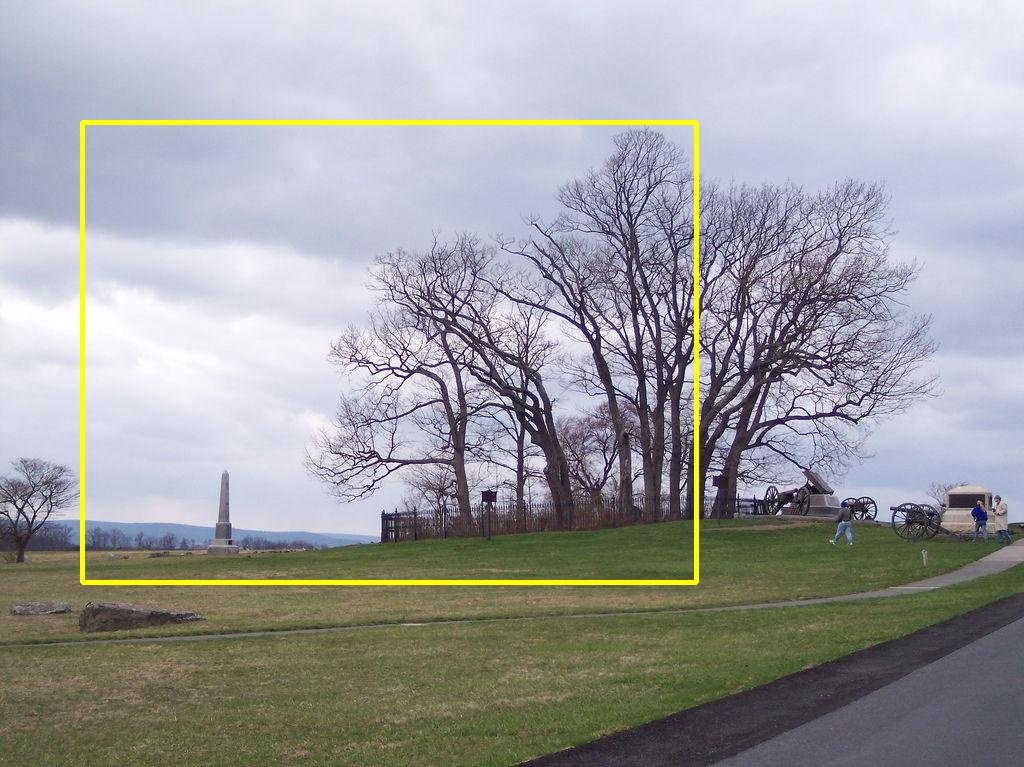}&
\includegraphics[width=3.2cm]{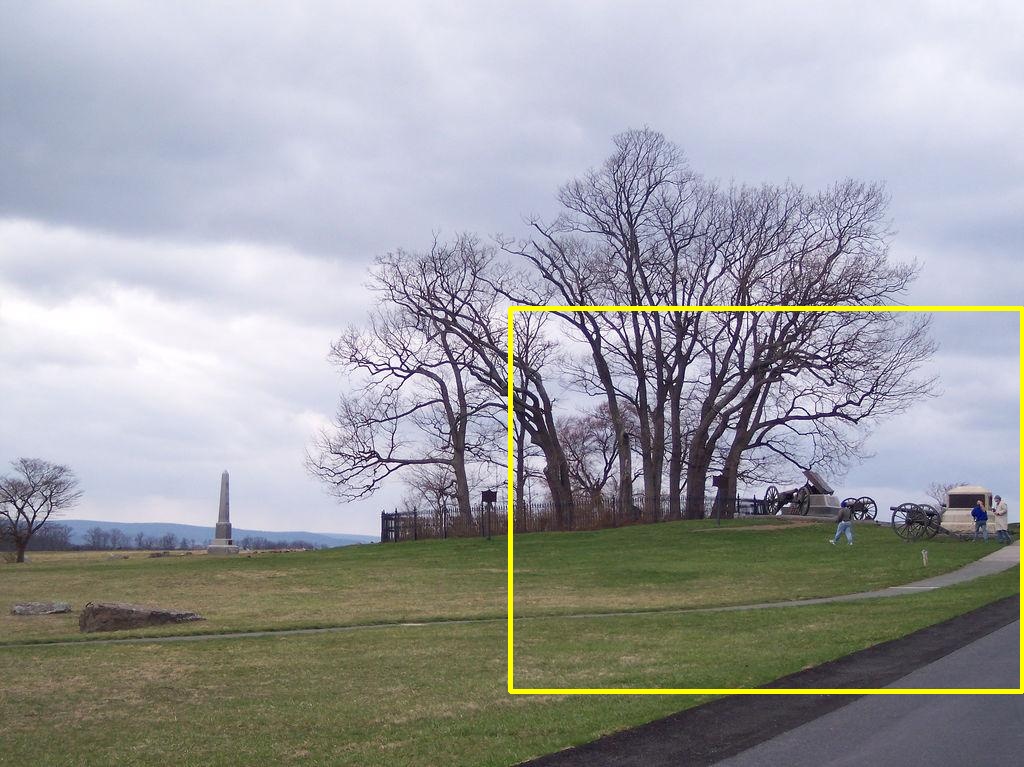}&
\includegraphics[width=3.2cm]{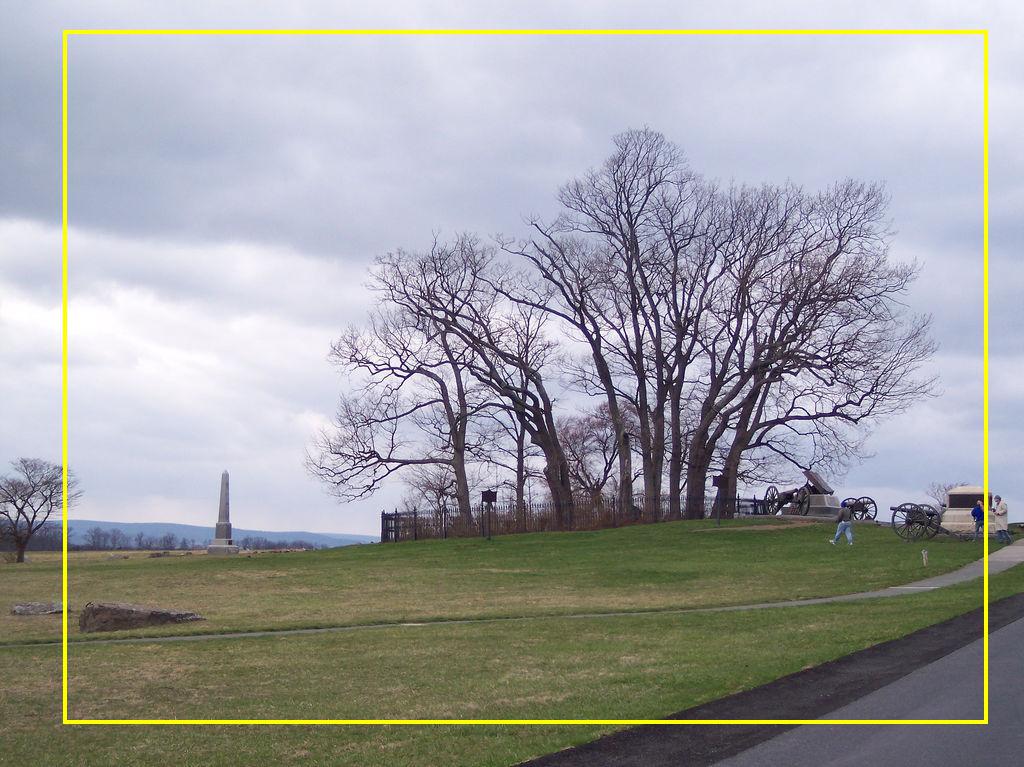}\\

\includegraphics[width=3.2cm]{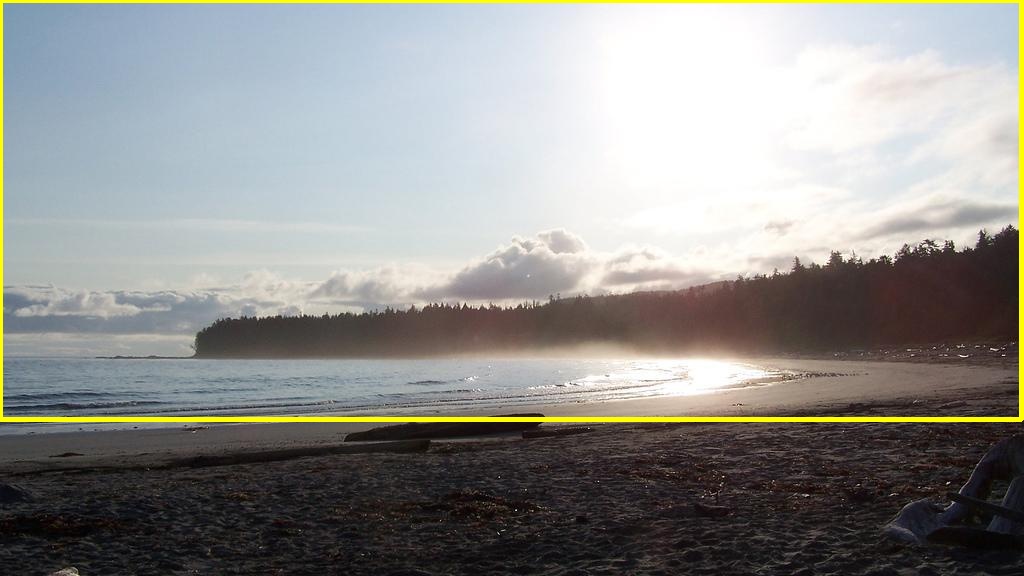}&
\includegraphics[width=3.2cm]{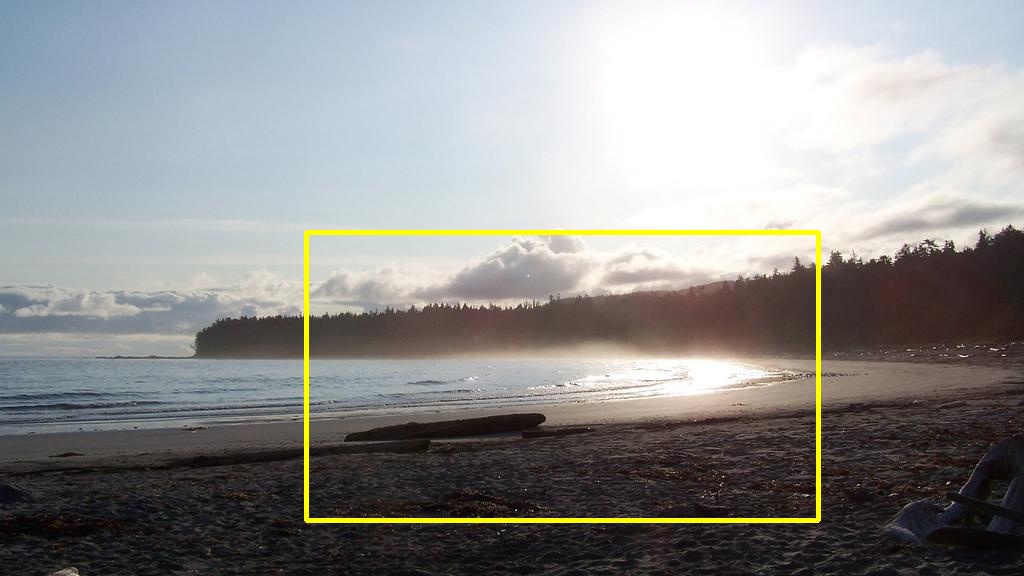}&
\includegraphics[width=3.2cm]{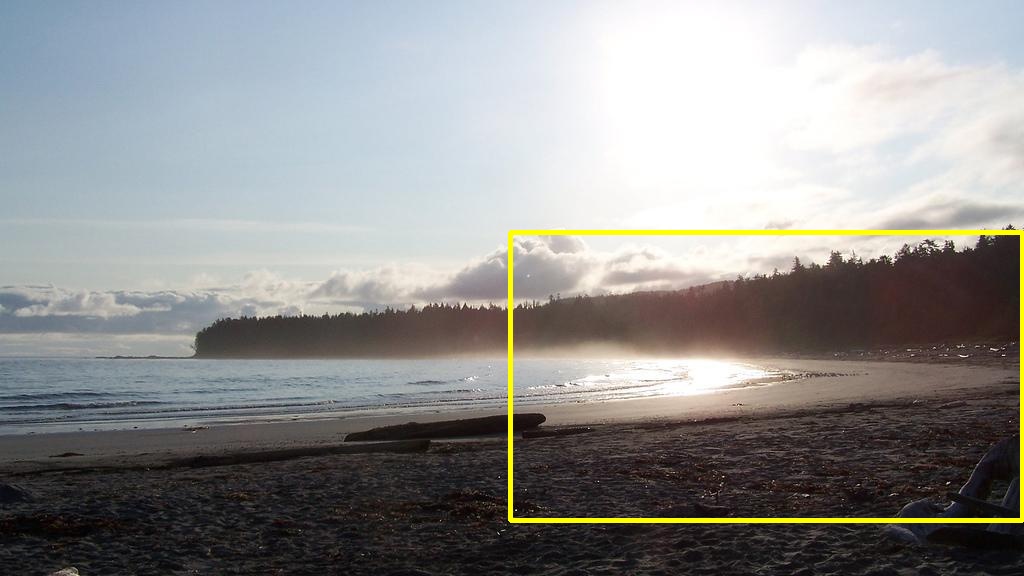}&
\includegraphics[width=3.2cm]{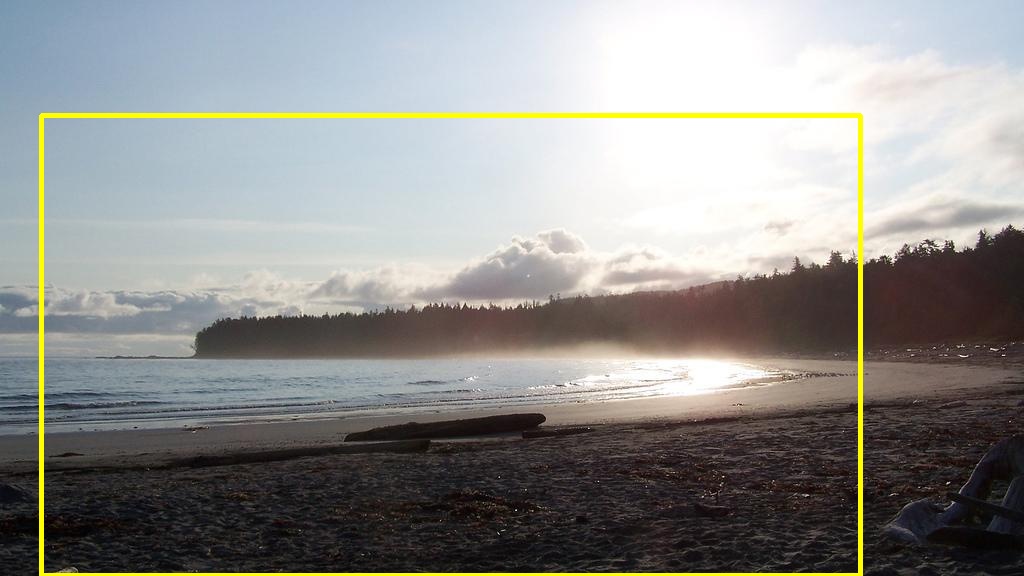}&
\includegraphics[width=3.2cm]{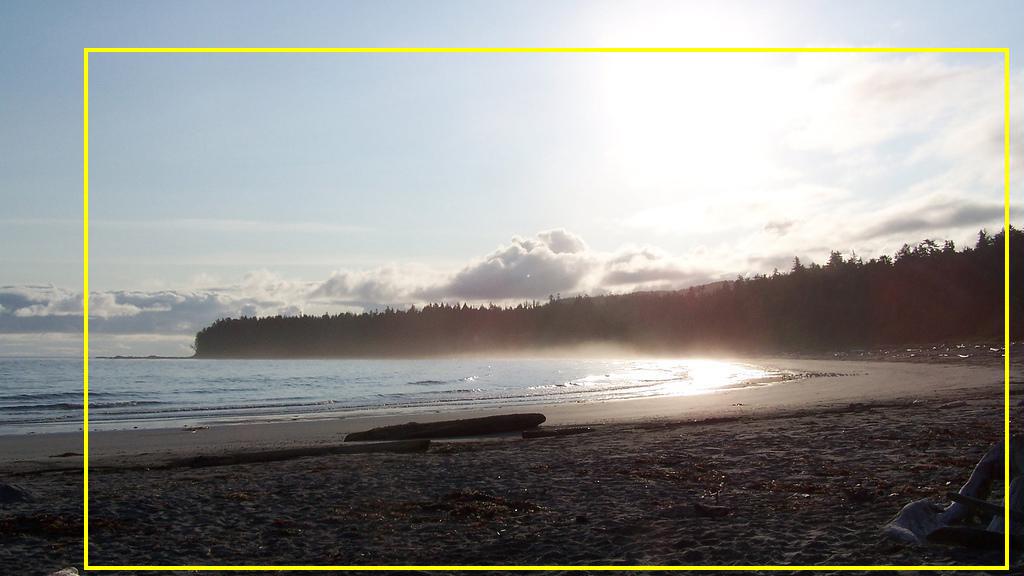}\\

\includegraphics[width=3.2cm]{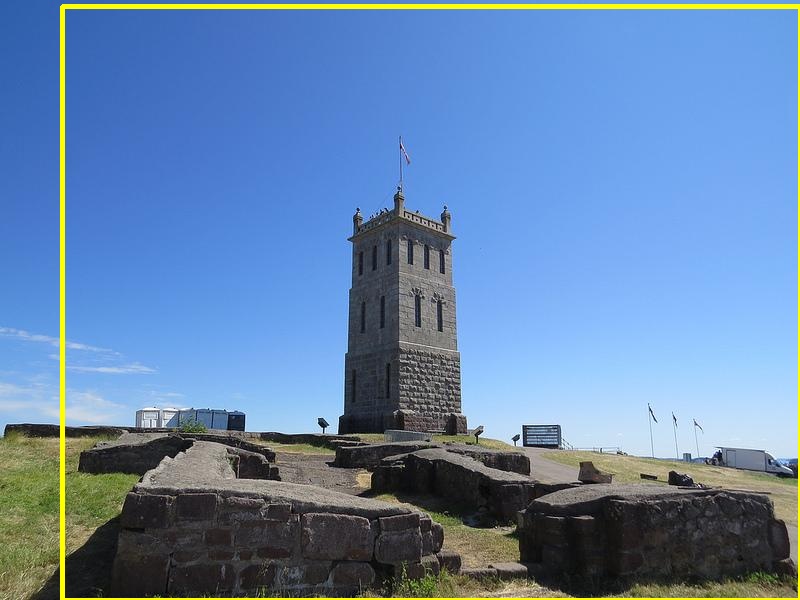}&
\includegraphics[width=3.2cm]{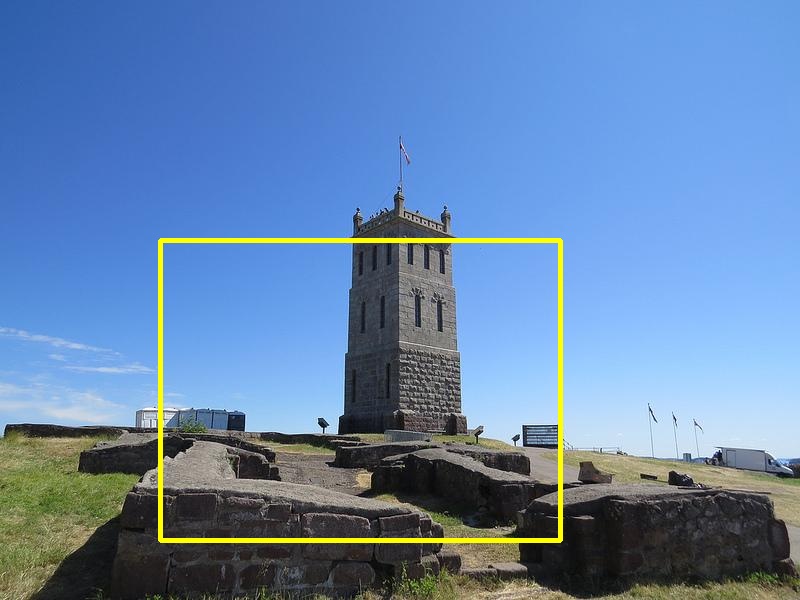}&
\includegraphics[width=3.2cm]{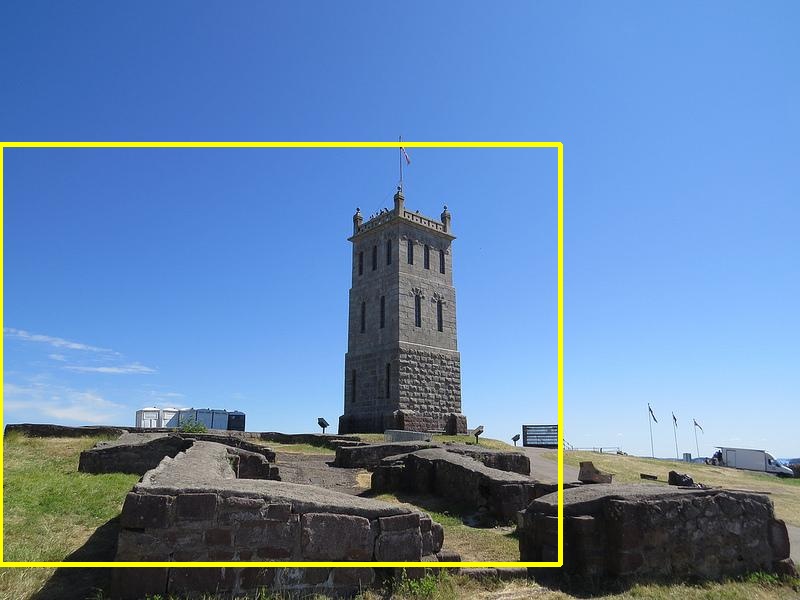}&
\includegraphics[width=3.2cm]{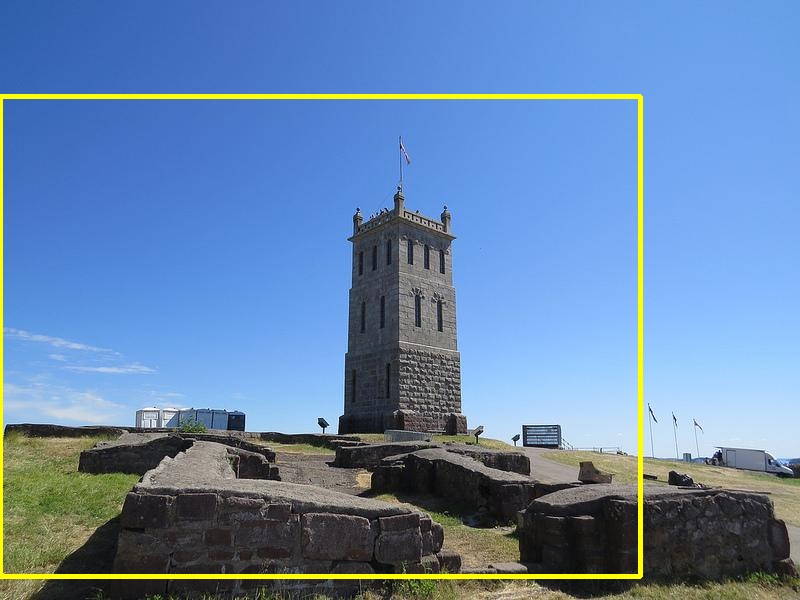}&
\includegraphics[width=3.2cm]{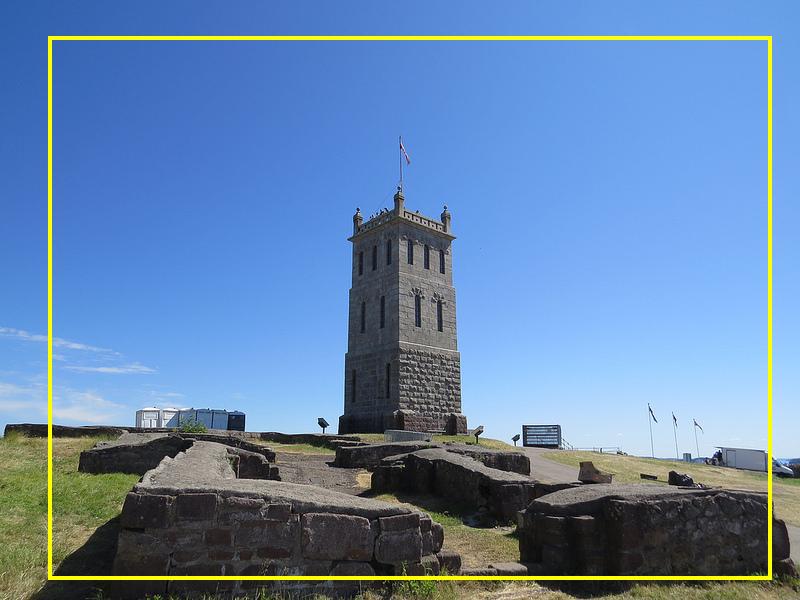}\\

\includegraphics[width=3.2cm]{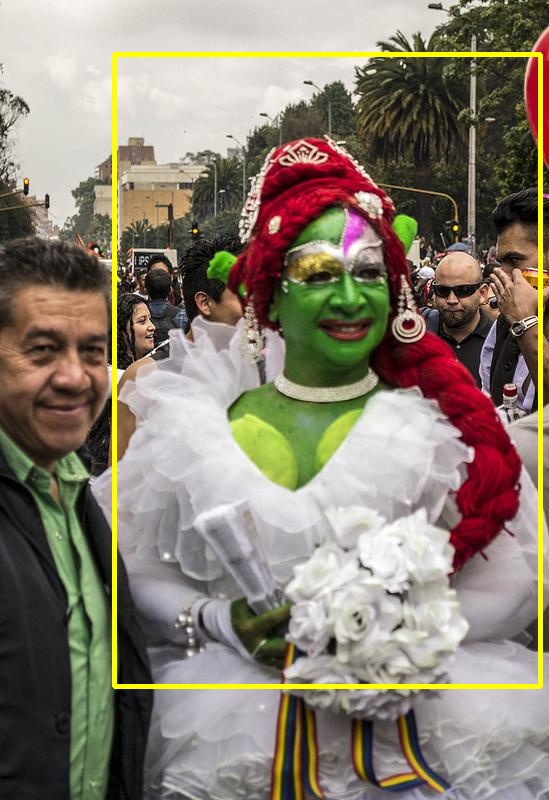}&
\includegraphics[width=3.2cm]{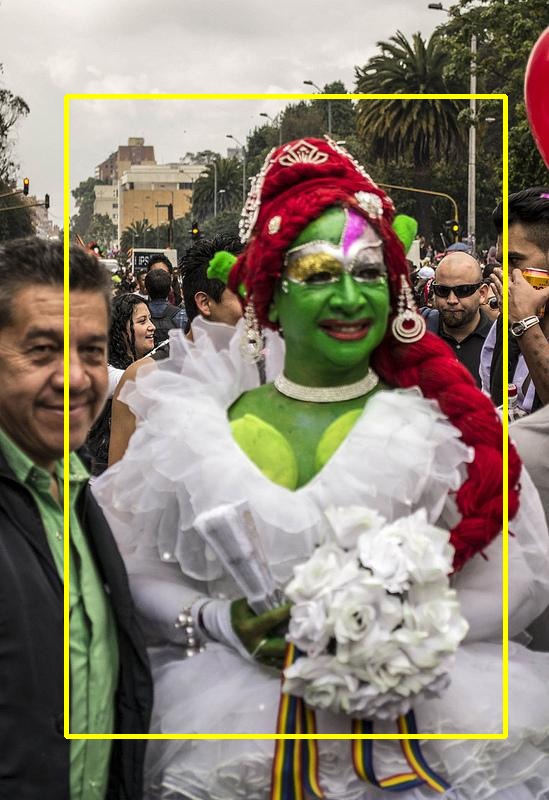}&
\includegraphics[width=3.2cm]{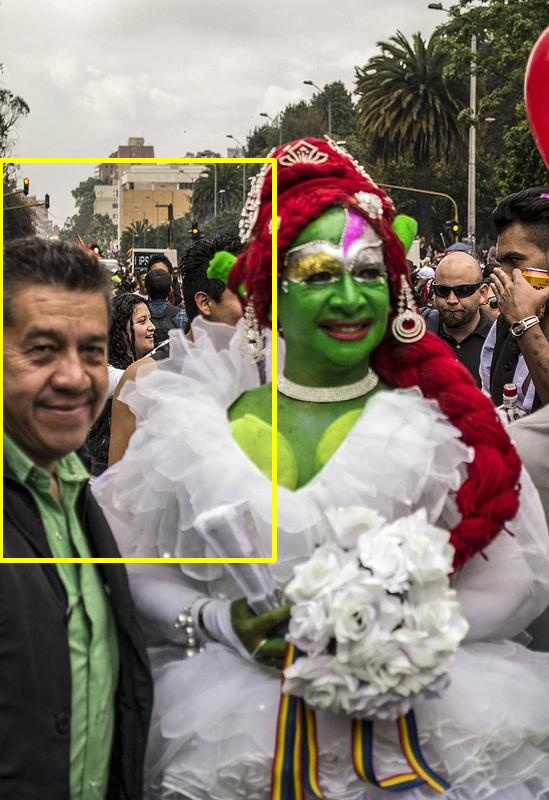}&
\includegraphics[width=3.2cm]{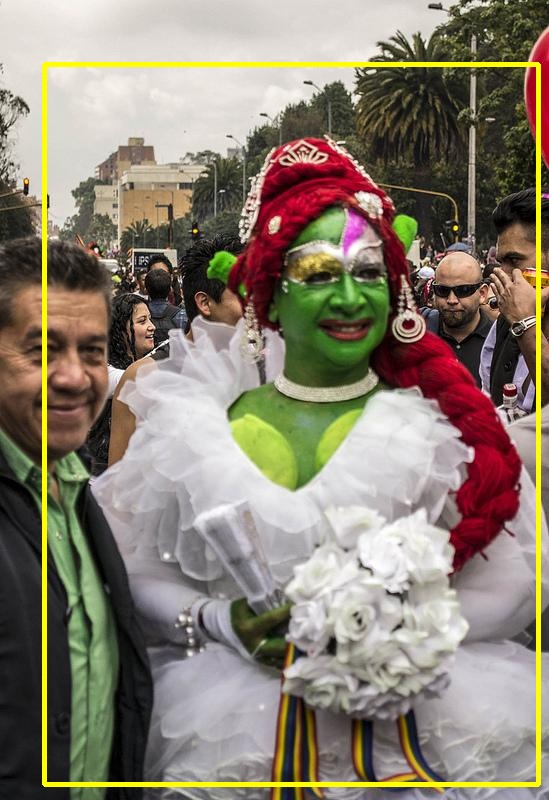}&
\includegraphics[width=3.2cm]{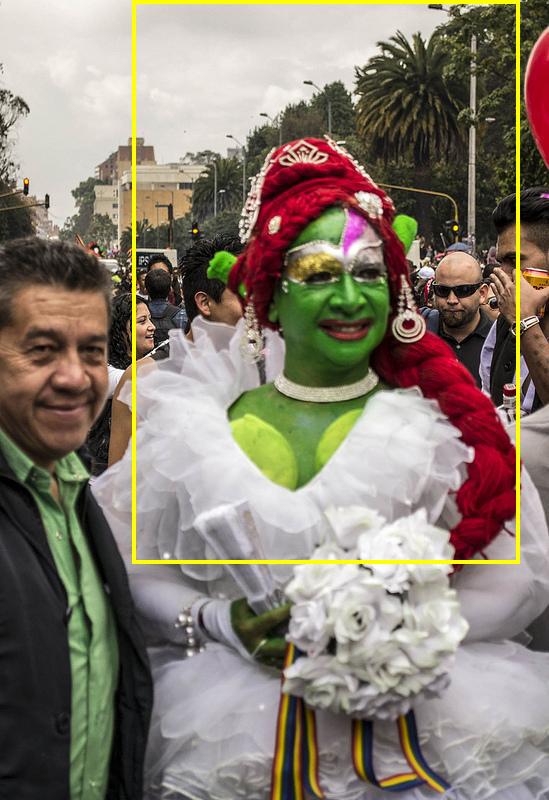}\\

(a) \small{Ground Truth} & (b) \small{eDN} & (c) \small{AlexNet\_finetue} & (d) \small{RankSVM+FCDB} & (e) \small{VFN}
\end{tabular}
\end{center}
\caption{Image cropping examples from FCDB \cite{Chen:WACV:2017}. The best crops determined by various methods are drawn as yellow rectangles. Best viewed in color.}
\label{fig:baseline_comparison}
\end{figure*}

Unlike traditional approaches, VFN learns to compose without explicitly modeling photo composition.
In a sense, it is accomplished by \emph{avoiding} the views violating photographic rules encoded in professional photographs.
Take the fifth row of \figname~\ref{fig:baseline_comparison} as an example, the baseline methods inappropriately cut through visually significant subjects.
Previous methods explicitly deal with such situation by modeling \emph{cut-through} feature \cite{Yan:CVPR:2013} or \emph{border simplicity} \cite{Lu:MM:2014}.
However, VFN naturally ignores these views because the proposed crop sampling method covers such cases and they are always penalized in our ranking model.
Due to the principle of pairing a good source image and a bad crop, there is thus the concern that the learned model is biased to favor larger views with more image content.
However, according to the benchmark, such tendency is not observed and the ranking model works well regardless of the scales.

Currently, VFN does not take full advantage of the SPP technique.
Its performance can potentially be further improved if the constraint of fixed-size input can be removed and the input images do not need to undergo undesired transformations (\eg, cropping or scaling) that typically cause damages to image composition \cite{Mai:CVPR:2016}.

We have shown that VFN is generic across categories in \secname~\ref{sec:performance_evaluation}.
It is considered that the generalization capability of VFN partially benefits from the object classification capability of the pre-trained AlexNet \cite{Krizhevsky:NIPS:2012}, which provides rich information to learn category-specific features that discriminate aesthetic relationships.

\paragraph{Limitations and Future Works}

The main limitation of VFN comes from its data sampling methodology, which only samples a sparse set of possible pairs of views.
The success of VFN can be accounted for that the aesthetic relations between the sampled pairs (\ie, a source image and a random crop) are definite.
However, it remains a challenging task for VFN to rank similar views whose aesthetic relation is ambiguous (\eg, two random crops or two nearly identical views).
Empirically, we found that evaluating a finer set of sliding windows with VFN causes the performance to degrade instead, which is possibly caused by the confusion between very similar views.
To maximize the performance of VFN, it is considered to incorporate a view selection procedure, which can effectively eliminate most unnecessary candidates and produces a sparse set of good candidates.
VFN currently needs to evaluate a number of proposal windows to accomplish view finding.
For future work, we plan to incorporate techniques like Faster R-CNN \cite{Ren:NIPS:2015} to improve its time efficiency.

\section{Conclusion}

In this work, we considered one of the most important problems in computational photography -- automatically finding a good photo composition.
Inspired by the thinking process of photo taking, 
a deep ranking network is proposed to learn the best photographic practices by leveraging human knowledge from the abundant professional photographs on the web.
We develop a costless and effective method to sample high-quality ranking samples in an unsupervised manner.
Without any hand-crafted features, the proposed method is simple and generic.
The resulting aesthetics-aware model is evaluated on two image cropping datasets and achieves state-of-the-art performance in terms of cropping accuracy.

{\small
\bibliographystyle{ieee}
\bibliography{egbib}
}

\end{document}